%% file: main.tex
\documentclass{article}

\PassOptionsToPackage{dvipsnames,table}{xcolor}
\usepackage[table]{xcolor}          
\usepackage{iclr2027_conference}

\makeatletter
\def\section{\@startsection{section}{1}{\z@}{-1.5ex plus -0.5ex minus -.2ex}%
  {1.0ex plus 0.3ex minus 0.2ex}{\large\sc\raggedright}}
\def\subsection{\@startsection{subsection}{2}{\z@}{-1.3ex plus -0.5ex minus -.2ex}%
  {0.5ex plus .2ex}{\normalsize\sc\raggedright}}
\def\paragraph{\@startsection{paragraph}{4}{\z@}{1.0ex plus 0.4ex minus .2ex}%
  {-1em}{\normalsize\bf}}
\makeatother
\usepackage{wrapfig}
\usepackage{amsmath}
\usepackage{amssymb}
\usepackage{booktabs}
\usepackage{graphicx}
\usepackage{microtype}
\usepackage{caption}
\renewcommand{\topfraction}{0.92}

\renewcommand{\textfraction}{0.06}
\renewcommand{\floatpagefraction}{0.75}

\usepackage[section]{placeins}  
\usepackage{float}              
\usepackage{enumitem}
\usepackage{wrapfig}             
\usepackage[most]{tcolorbox}     

\usepackage[T1]{fontenc}
\usepackage{newtxtext,newtxmath}
\definecolor{oursemph}{RGB}{0,90,181}   
\definecolor{failred}{RGB}{178,24,43}   
\definecolor{accentcyan}{HTML}{0B828D}   
\definecolor{newres}{RGB}{223,245,223}  
\definecolor{notmeasured}{RGB}{178,24,43}  
\definecolor{finalres}{RGB}{247,233,178}  
\newcommand{\important}[1]{\textit{\textcolor{accentcyan}{#1}}}

\usepackage[breaklinks,colorlinks,linkcolor=failred,citecolor=oursemph,urlcolor=oursemph]{hyperref}
\usepackage{url}

\newtcolorbox{promptbox}[1]{%
  enhanced, breakable, colback=gray!8, colframe=gray!40, boxrule=0.4pt,
  coltitle=black, colbacktitle=gray!25,
  fonttitle=\bfseries\small, fontupper=\ttfamily\small,
  title=#1, sharp corners, left=6pt, right=6pt, top=4pt, bottom=4pt,
}

\usepackage{pifont}
\definecolor{cspromptbg}{HTML}{EAF2FC}
\definecolor{cspromptfr}{HTML}{9EC2E6}
\definecolor{csbarplain}{HTML}{F0F0F0}
\definecolor{csbarhi}{HTML}{DCE9F8}
\definecolor{csframe}{HTML}{B4B4B4}
\definecolor{csframehi}{HTML}{2A6BB0}
\definecolor{csokc}{HTML}{0B7A3D}
\definecolor{csbadc}{HTML}{B2182B}

\newtcolorbox{caseprompt}{%
  enhanced, arc=2pt, boxrule=0.5pt,
  colback=cspromptbg, colframe=cspromptfr,
  left=6pt, right=6pt, top=4pt, bottom=4pt,
  fontupper=\small,
}

\newtcolorbox{casepanel}[4]{%
  enhanced, arc=2pt, boxrule=0.5pt,
  colback=white, colframe=#2,
  colbacktitle=#1, coltitle=black,
  fonttitle=\small, title={#3\hfill #4},
  left=6pt, right=6pt, top=4pt, bottom=4pt,
  fontupper=\footnotesize\ttfamily,
  before upper=\raggedright,
}

\newcommand{\csok}{\textcolor{csokc}{\ding{51}\,\textbf{Correct}}}

\newcommand{\csmodel}[2]{\textbf{#1}\hspace{0.5em}{\footnotesize\color{black!55}(#2)}}
\newcommand{\csnum}[1]{\textcolor{oursemph}{#1}}
\newcommand{\cserr}[1]{\textcolor{csbadc}{#1}}
\newcommand{\csdisp}[1]{\par\vspace{3pt}\centerline{#1}\vspace{3pt}}
\newcommand{\csnote}[1]{\par\vspace{3pt}{\normalfont\itshape\footnotesize #1}}
\newcommand{\csgap}{\\[4pt]}

\title{Commit Locally, Exit Globally:\\Coordinating Adaptive Sampling and Early Exit in Diffusion Language Models}

\author{%
Chia-Ming Lee$^{1,2}$,\;
Shao-Kai Liu$^{1}$,\;
Ming-Ching Chang$^{2}$,\;
Xin Li$^{2}$\\[2pt]
\bf Yu-Lun Liu$^{1}$,\;
Chih-Chung Hsu$^{1}$\\[4pt]
$^{1}$National Yang Ming Chiao Tung University \quad
$^{2}$University at Albany, SUNY}

\iclrfinalcopy  

\begin{document}

\maketitle
\lhead{Preprint}  
\vspace{-2mm}
\begin{abstract}\vspace{-2mm}
Diffusion language models predict tokens in parallel but still require many iterative forward
passes. Adaptive sampling accelerates local token commitment, while recent global stopping
methods seek to end generation early. Our trajectory analysis shows that short answers can stabilize early in decoding,
whereas reasoning outputs often require more steps to settle. A fixed progress threshold
therefore struggles to avoid both redundant computation on short-answer tasks and premature
termination on reasoning tasks.
We introduce \textbf{$\mathrm{C}^4$}, a training-free framework that combines local decoding
acceleration with adaptive global early exit.
\textbf{Confidence-Verified Early Exit} (CVEE) determines when to end generation,
skipping the remaining steps once its stopping criterion is met. It combines confidence
and temporal agreement to adapt stopping to the observed revision history.
\textbf{Commit-Core-Then-Confirm} (CCTC) accelerates token commitment while generation
continues, reducing the decoding steps needed to complete each block. It extends
confidence-threshold parallel decoding with a boundary-anchored core and one-step
confirmation of deferred predictions, constraining additional commitments to mitigate
premature freezing under unresolved context.
With CVEE calibrated once on LLaDA, the same numerical settings transfer across all $12$
zero-shot tasks and to Dream without recalibration. Under this configuration, $\mathrm{C}^4$ reduces
decoding steps by $64$--$95\%$ and achieves measured decoding-loop speedups of
$2.6\times$--$18.6\times$, with task performance close to full decoding. Component analyses show complementary contributions: early exit supplies
most savings on short-answer tasks, while local acceleration contributes more on reasoning
and code generation. These results demonstrate that $\mathrm{C}^4$ effectively combines local token commitment and global early exit to accelerate diffusion decoding while largely preserving task performance. Project page: \url{https://ming053l.github.io/c4-dllm.github.io/}.
\end{abstract}\vspace{-2mm}

\FloatBarrier

\input{sections/introduction}

\input{sections/related_work}

\input{sections/preliminaries}

\input{sections/method}

\input{sections/experiments}

\input{sections/conclusion}

\subsection*{AI use statement}
Generative AI tools assisted with code implementation and debugging, result aggregation,
figure preparation, manuscript drafting and editing, and literature and bibliographic checks.
They also supported consistency checks across the method description, reported results, and
released code, informing revisions to empirical claims. The authors reviewed and verified the
resulting code, analyses, references, and text and take full responsibility for the final content.

\bibliographystyle{iclr2027_conference}
\bibliography{main}

\newpage
\appendix
\input{sections/appendix/overview}

\input{sections/appendix/hyperparams}

\input{sections/appendix/retiming}

\input{sections/appendix/span_control}

\input{sections/appendix/cvee_trace}

\let\savedFloatBarrier\FloatBarrier
\let\FloatBarrier\relax
\input{sections/appendix/wino_cvee}
\let\FloatBarrier\savedFloatBarrier

\input{sections/appendix/taublk}

\input{sections/appendix/scaling_cache}

\input{sections/appendix/fill_quality}

\input{sections/appendix/limitations}

\end{document}

%% file: sections/introduction.tex
\vspace{-3mm}\section{Introduction}
\label{sec:intro}
\vspace{-1mm}
Diffusion language models (DLMs) use bidirectional context to predict masked tokens in
parallel \citep{nie2025llada,ye2025dream}. In block-wise decoding, they generate a sequence
one block at a time, resolving multiple token positions within each block through iterative
parallel updates. This parallelism offers the potential for faster generation than the
one-token-at-a-time decoding of autoregressive language models. Realizing this potential
requires deciding both how quickly to resolve each block and how long to continue generation.
Inference can therefore be accelerated at two scopes: \emph{locally}, by committing tokens
sooner and advancing to the next block, and \emph{globally}, by terminating the sequence before
its decoding budget is exhausted.

Specifically, tasks differ in where these opportunities arise: short answers can stabilize early, allowing
later blocks to be skipped, whereas extended reasoning often benefits more from faster updates
within blocks as predictions continue evolving. Decision criteria should therefore reflect both these
dynamics and the action's scope: token-level confidence can guide local commitment, while
global stopping requires evidence that the task-relevant prediction has stabilized.

Existing methods typically exploit one of these opportunities. At the local level,
adaptive samplers such as Fast-dLLM \citep{wu2025fastdllm}, SlowFast Sampling
\citep{wei2025slowfast}, and KLASS \citep{kim2025klass} accelerate token commitment;
WINO \citep{hong2026wino} additionally permits revocation through re-masking.
By completing blocks faster, these methods reduce the cost of continued generation, but
leave open when the remaining blocks can be skipped altogether. Recent work on global
stopping begins to address this question: Prophet \citep{li2026diffusion} and SchED
\citep{mohamed2025sched} use region-level confidence or schedule-dependent rules to
terminate generation early. Their savings, however, depend on how much computation remains
at the stopping point, while preceding within-block updates still rely on the underlying
sampler. When predictions settle late, global stopping alone therefore leaves limited room
for acceleration, making local sampling efficiency especially important for reducing the cost
of the blocks that must still be generated before termination.

Combining these complementary sources of savings calls for a stopping rule that adapts to
observed decoding trajectories across tasks and backbones while keeping its numerical settings
fixed. The trajectory itself, however, also depends on local sampling: committing tokens
reshapes the context for subsequent predictions and thus the confidence and agreement signals
available for stopping. Local acceleration therefore changes not only how much computation
remains, but also the evidence used to decide whether that computation is still needed.
This connection motivates designing local commitment and global stopping together around
the same evolving generation.

We introduce \important{$\mathrm{C}^4$} around this division of responsibilities.
\important{Confidence-Verified Early Exit (CVEE)} combines current confidence with temporal
agreement, adapting its patience to the history of revisions. \important{Commit-Core-Then-Confirm
(CCTC)} extends confidence-threshold parallel decoding with a boundary-anchored core and
one-step confirmation of deferred predictions. It retains the reference decoder's scheduled
commitments and operates alongside CVEE on the same trajectory. Together, the components target
two sources of excess computation: repeated steps within a block and continued decoding after
stabilization. Moreover, because CVEE adjusts its patience to observed revisions rather than a fixed
decoding-progress threshold, it can accommodate different stabilization times without changing
its numerical settings, reducing reliance on task-specific stopping calibration.
\vspace{-1mm}
\begin{itemize}[leftmargin=*,itemsep=2pt,topsep=4pt]
\vspace{-1mm}
\item We analyze stabilization timing across diffusion decoding trajectories, identifying
how early-exit opportunities vary between short-answer and long-reasoning tasks. This analysis
motivates confidence and temporal agreement as complementary signals for global stopping.
\vspace{-1mm}
\item We introduce \important{$\mathrm{C}^4$}, a training-free framework that combines local
acceleration and global termination with reduced reliance on task-specific calibration.
Because local commitment reshapes the trajectory that stopping observes, their savings are not additive; CVEE and CCTC are designed to act on this shared trajectory.
\vspace{-1mm}
\item Across $12$ zero-shot tasks on LLaDA and Dream, $\mathrm{C}^4$ achieves
$2.6\times$--$18.6\times$ measured decoding-loop speedups while largely preserving task
performance. Our experiments show how its complementary components sustain savings across
tasks: global early exit skips computation after early stabilization, while local commitment
accelerates the blocks needed for extended reasoning and code generation.

\end{itemize}

%% file: sections/related_work.tex
\vspace{-5mm}
\section{Related Work}
\label{sec:related}\vspace{-1mm}

\noindent\textbf{Diffusion language models.}
The diffusion framework traces to \citet{sohldickstein2015diffusion}, with extensions to discrete
data by \citet{hoogeboom2021argmax,austin2021d3pm,campbell2022continuous}.
\citet{lou2023sedd} and the masked-diffusion line
\citep{shi2024mdlm,sahoo2024mdlm,ou2024radd,zheng2024masked} establish the parameterizations
underlying modern DLMs. LLaDA \citep{nie2025llada} and Dream \citep{ye2025dream}, the open-weight
models we evaluate, support parallel prediction at masked positions with flexible inference
budgets. Under block-wise decoding, this flexibility creates two opportunities to reduce
computation: completing individual blocks in fewer steps and skipping the remaining sequence.

\vspace{-1mm}
\noindent\textbf{Adaptive sampling for DLMs.}
Adaptive sampling controls which token positions commit while decoding continues.
Fast-dLLM \citep{wu2025fastdllm}, SlowFast Sampling \citep{wei2025slowfast}, and KLASS
\citep{kim2025klass} commit using token-level statistics, while concurrent work refines related
signals: LESS \citep{mohamed2026less} combines confidence with stability, STDec \citep{chen2026stdec} adapts thresholds using neighboring decoded states and temporal stability, TACG
\citep{wang2026tacg} uses EMA-logit trajectories, Swordsman \citep{zhang2026swordsman}
re-partitions blocks along entropy shifts, $R^2$-dLLM \citep{du2026r2dllm} exploits
spatio-temporal redundancy through decoding rules and fine-tuning, and \citet{kim2026earlydecisions} identify proximity
bias in the denoising order. WINO \citep{hong2026wino}
permits broad commitment followed by re-masking of disputed positions. CCTC builds on
confidence-threshold parallel decoding by structuring additional commitments around a
boundary-anchored core and confirming deferred predictions at the next step. It retains
scheduled commitments and does not revoke committed tokens. This local control reduces
within-block computation while leaving sequence termination to CVEE.

\vspace{-1mm}
\noindent\textbf{Early exit and sequence termination.}
Early exit has been studied at the layer, reasoning-step, and token levels.
\citet{gu2026semanticfixed} apply it to \emph{depth} via hidden-state stabilization.
S-GRPO \citep{dai2025sgrpo} and DEER \citep{yang2026deer} study early exit in reasoning
models. Related diffusion methods address revision and verification: CORE \citep{zhai2026core}
identifies tokens sensitive to masked-context perturbations, while BMC \citep{bmc2026manifold}
evaluates consistency through masking and reconstruction. For DLMs, Prophet \citep{li2026diffusion} and SchED
\citep{mohamed2025sched} stop using aggregate confidence signals over fixed monitoring regions or
schedule-dependent thresholds to determine when generation should end.
CVEE bases this decision on confidence and temporal agreement of the task-relevant prediction,
with patience adapted to its revision history. This shifts the stopping criterion from decoding
progress to observed prediction behavior, allowing numerical settings to be reused across
the evaluated tasks and backbones. Together with CCTC, it addresses both the cost of ongoing
generation and the point at which further decoding can be skipped.

%% file: sections/preliminaries.tex
\vspace{-2mm}\section{Preliminaries}
\label{sec:dlm-prelim}\vspace{-2mm}

We describe the masked generation process and the block-wise schedule used in our evaluation,
then distinguish the local commitment and global stopping decisions that control its cost.

\vspace{-1mm}
\noindent\textbf{Masked diffusion generation.}
A DLM generates a sequence $x_0$ of length $L$ by reversing a discrete corruption process that
progressively masks a clean sequence over forward time $u\in[0,1]$,\vspace{-1mm}
\begin{equation}
q(x_u \mid x_0) = \textstyle\prod_{i=1}^{L} q(x_u^i \mid x_0^i),
\label{eq:forward-process}
\end{equation}\vspace{-2mm}
\begin{equation}
q(x_u^i \mid x_0^i) = \left\{\begin{array}{ll@{\quad}l}
x_0^i, & \text{w.p. } 1-u, & \text{(a) token kept} \\
\texttt{[MASK]}, & \text{w.p. } u, & \text{(b) token masked}
\end{array}\right.
\label{eq:mask-kernel}
\end{equation}
so $x_1$ is fully masked and $x_0$ is the original sequence.
The model $p_\theta(x_0\mid x_u)$ learns to recover the clean sequence from masked inputs.
Generation starts from masks and proceeds through $T$ discrete decoding steps, indexed by
$t\in\{1,\ldots,T\}$. At each step, the model predicts masked positions in parallel, and a
remasking rule selects predictions to commit. The remaining positions stay masked for the
next step. We use $u$ for forward corruption time and $t$ for the decoding-step index.

\vspace{-1mm}
\noindent\textbf{Block-wise decoding.}
We partition the generation region into $N$ contiguous blocks $B_0,\ldots,B_{N-1}$.
Blocks are processed from left to right, with parallel prediction inside the active block and
$S=T/N$ scheduled steps per block. We use this semi-autoregressive schedule for both
LLaDA-8B-Instruct \citep{nie2025llada} and Dream-7B-Instruct \citep{ye2025dream}.
At each step, low-confidence remasking commits the highest-confidence predictions among
currently masked positions. The reference schedule sets the number of commitments so that
all positions are filled within the block budget. Local acceleration can select additional positions
to commit, allowing a completed block to advance before that budget is exhausted. Thus,
the schedule specifies a decoding budget, while the commitment rule determines how many
of those steps are actually needed to complete each block.

\vspace{-1mm}
\noindent\textbf{Sequence-level stopping.}
Whereas local commitment advances the active block, a sequence-level stopping gate decides
whether to end the entire generation, skipping all subsequent decoding steps.
For example, Prophet \citep{li2026diffusion} computes the mean top-1/top-2 logit gap over
a monitored region $\mathcal{R}$ to summarize the separation between the two leading predictions
at each position,\vspace{-1mm}
\begin{equation}
\bar g_t =
\tfrac{1}{|\mathcal{R}|}\textstyle\sum_{i\in\mathcal{R}}
\left(\ell_i^{(t),1}-\ell_i^{(t),2}\right),
\label{eq:prophet-gap}
\end{equation}
where $\ell_i^{(t),1}$ and $\ell_i^{(t),2}$ are the two largest logits at position $i$ and step $t$.
We use a fixed task-format-specific definition of $\mathcal{R}$ in our evaluation, as detailed in
Appendix~\ref{app:hyperparams-searchmode}. Prophet fills all remaining masked positions once
$\bar g_t$ crosses its published progress-dependent three-stage threshold; SchED replaces this
with a progress-dependent decay schedule and a stability guard \citep{mohamed2025sched}.
Local commitment and global stopping therefore act on the same evolving prediction but
control different portions of the computation. In the following section, we examine how
prediction stabilization informs stopping and develop the two controls within a shared decoder.

%% file: sections/method.tex
\vspace{-2mm}\section{Local Commitment and Global Exit}
\label{sec:method}\vspace{-1mm}

\begin{figure}[!t]
\centering\vspace{-3mm}
\includegraphics[width=\linewidth]{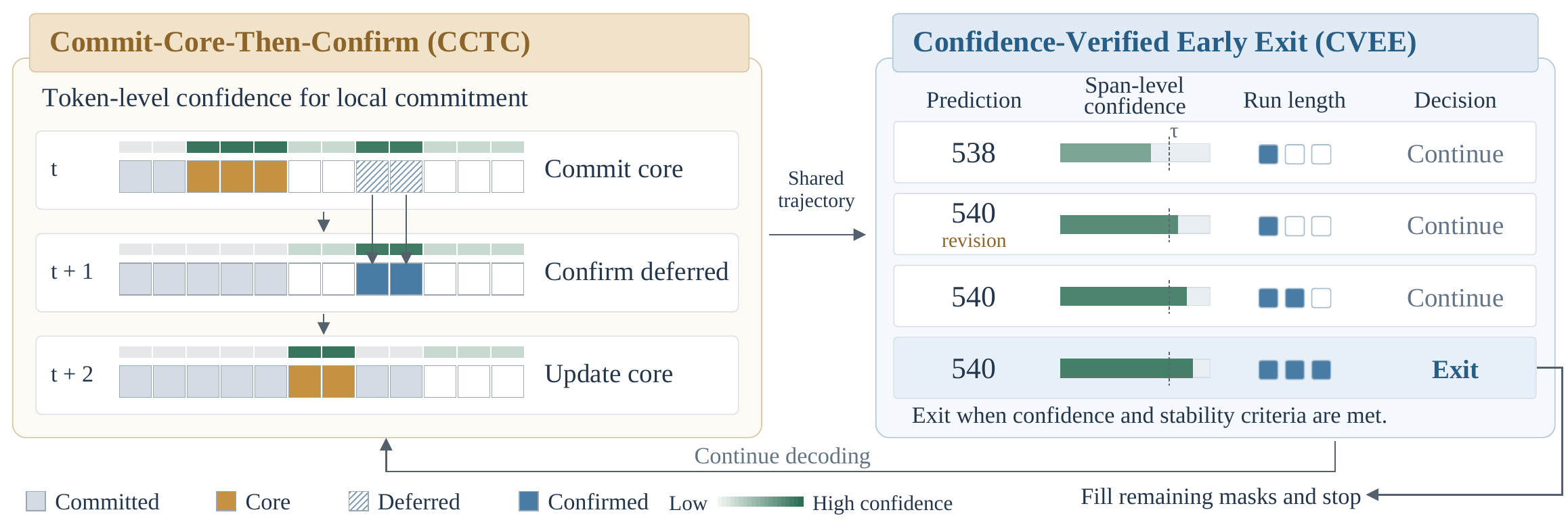}
\caption{\textbf{$\mathrm{C}^4$ accelerates decoding at two complementary scales.}
CCTC uses token-level confidence to advance a boundary-anchored core and confirm deferred
predictions as context develops. Along this shared trajectory, CVEE combines mean span-level
confidence with temporal agreement to stop generation. Local commitment saves computation
within blocks; global stopping removes later steps.}\vspace{-1mm}
\label{fig:teaser}
\end{figure}

$\mathrm{C}^4$ reduces decoding cost through two decisions made along the same trajectory:
which tokens to commit within the active block, and when to end the sequence.
\important{CCTC} accelerates block completion while generation continues;
\important{CVEE} skips the remaining steps when confidence and temporal agreement support
termination (Figure~\ref{fig:teaser}). We first examine when predictions stabilize, then develop
the global stopping rule and the local commitment mechanism.

\vspace{-2mm}
\subsection{Stabilization Timing and the Opportunity for Early Exit}
\label{sec:convergence-floor}\vspace{-1mm}

To determine when global stopping can save computation, we first examine when predictions
settle during decoding. Full decoding trajectories let us compare this timing across tasks
and establish the behavior that an adaptive stopping rule needs to accommodate.

\vspace{-1mm}
\noindent\textbf{Measuring stabilization along decoding trajectories.}
We measure when extracted answers stop changing in full LLaDA-8B-Instruct decodes drawn
equally from MMLU, GSM8K, and MATH. Let $a_{j,r}$ denote the extracted candidate from trajectory
$j$ at normalized decoding progress $r$, and let $a_j^{\mathrm{full}}$ be its full-decoding answer.
We define\vspace{-1mm}
\[
s_j = \min\{s : a_{j,r}=a_j^{\mathrm{full}}\ \forall r\geq s\}
\]
as trajectory $j$'s \emph{stabilization progress}, and $s_p$ as the progress by which a fraction
$p$ of trajectories have reached their own $s_j$. This retrospective measure identifies when the extracted answer reaches
its final value, irrespective of whether that value is correct.

\vspace{-1mm}
\noindent\textbf{Different tasks settle at different stages.}
Using this measure, we find that $s_{0.9}$ ranges from $0.04$ on short-answer tasks to $0.96$
on long-reasoning ones. Early stabilization leaves substantial computation available to skip,
whereas later stabilization leaves less room for early exit and makes within-block acceleration
more important. Consequently, a fixed stopping point cannot accommodate both patterns: waiting longer
forgoes savings on early-settling trajectories, while stopping sooner can interrupt predictions
that are still changing. We therefore use observed confidence and temporal agreement to
adapt the stopping decision to each trajectory.

\vspace{-2mm}
\subsection{Confidence-Verified Early Exit: Deciding When to Stop}
\label{sec:analysis}\vspace{-1mm}

\begin{figure}[t]
\centering\vspace{-3mm}
\includegraphics[width=\linewidth]{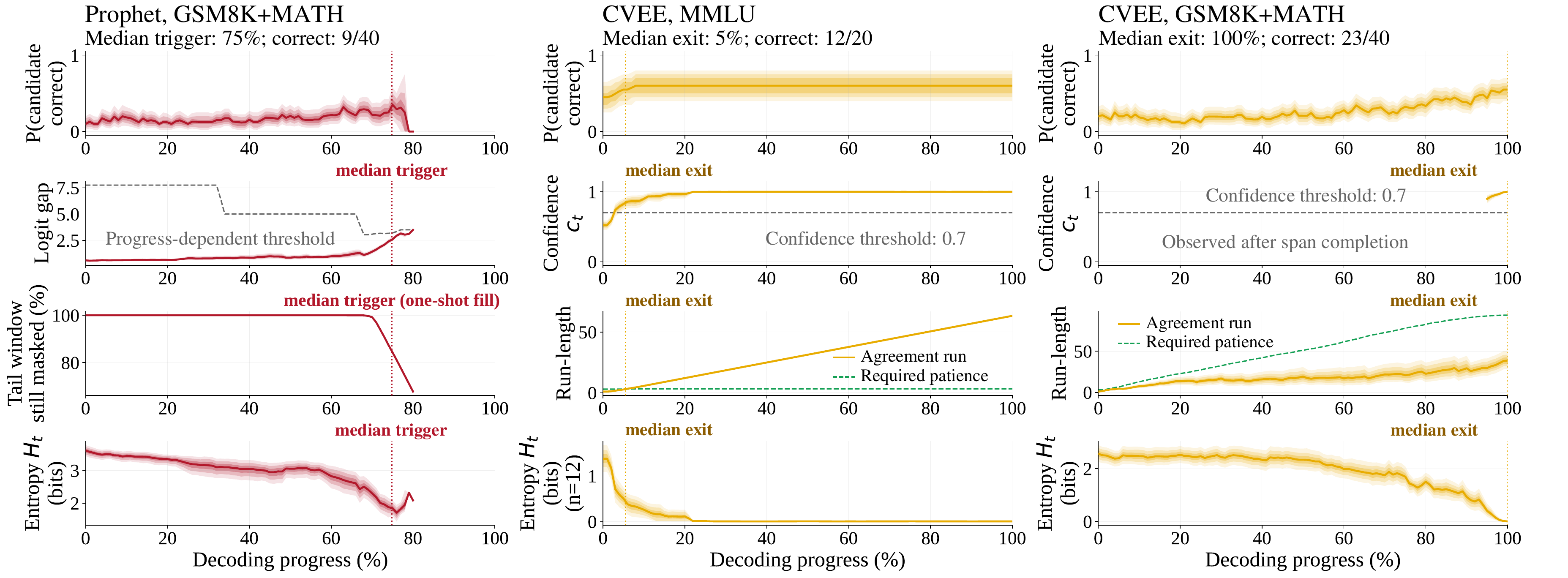}
\caption{\textbf{Stabilization timing shapes the opportunity for early exit.}
MMLU predictions settle early, leaving much of the decoding budget unused, whereas GSM8K
and MATH predictions continue to change later in decoding. CVEE exits early on MMLU and
waits longer on the reasoning tasks; Prophet can trigger while those predictions are still
changing. These trajectories illustrate why stopping should track evolving predictions
rather than elapsed decoding progress alone.}\vspace{-2mm}
\label{fig:cvee-trace}
\end{figure}

The retrospective analysis above uses the final output to identify stabilization. During
decoding, however, that output is not yet available. CVEE therefore combines current confidence with agreement across decoding steps and increases patience
when predictions undergo more revisions. This lets stopping respond to the evolving output
rather than to a fixed fraction of the decoding budget (Figure~\ref{fig:cvee-trace}).

\vspace{-1mm}
\noindent\textbf{Tracking confidence and temporal agreement.}
To track the relevant prediction as its location changes, at step $t$ a deterministic parser extracts and normalizes an answer from the current prediction
$\hat{x}^{(t)}$, relocating its span as decoding proceeds. We denote this extracted value by $a_t$. CVEE applies to outputs with a consistently extractable answer unit.
Conceptually, confidence describes the current prediction, while temporal agreement records
its persistence across steps. Their combination provides the evidence for the stopping rule below.

\vspace{-1mm}
\noindent\textbf{Revision-adaptive patience.}
Let $\mathrm{run}_t$ be the number of successive same-value
extractions (paused when extraction is unavailable); $\mathrm{changes}_t$ the number of changes in $a_t$;
and $c_t$ the mean current top-1 probability at positions in the eligible extracted span.
CVEE fires at the first step satisfying\vspace{-1mm}
\begin{align}
c_t &\geq \tau_{\text{CVEE}}
&&\text{(a) confidence gate} \nonumber\\
\mathrm{run}_t
&\geq \max\big(p_{\min},\,\lceil\gamma\cdot\mathrm{changes}_t\rceil\big)
&&\text{(b) stability gate}.
\label{eq:gate}
\end{align}
Here $p_{\min}$ sets minimum patience, while $\gamma$ increases the required waiting period
after revisions. Because this requirement follows the observed history, the gate has no explicit dependence
on normalized decoding progress or a prescribed stopping step. Patience adapts to observed
revisions, allowing the same numerical criterion to be used across decoding budgets and task
regimes; we calibrate it once and freeze it across tasks and backbones (Section~\ref{sec:setup}).
The block schedule nevertheless determines observation eligibility: spans lie within active or previously
decoded blocks. Because extraction retains committed tokens, which may differ from current top-1 predictions, CVEE is an empirical stopping criterion rather than a verification of the retained answer. Once the gate fires, the decoder fills remaining
masks from current predictions. Appendix~\ref{app:hyperparams-searchmode} details the updates.

\vspace{-2mm}
\subsection{Commit-Core-Then-Confirm: Selecting Parallel Commitments}
\label{sec:blockaccel}\vspace{-1mm}
\label{sec:sgpd}

\begin{figure*}[!t]
\centering\vspace{-1mm}
\includegraphics[width=\textwidth]{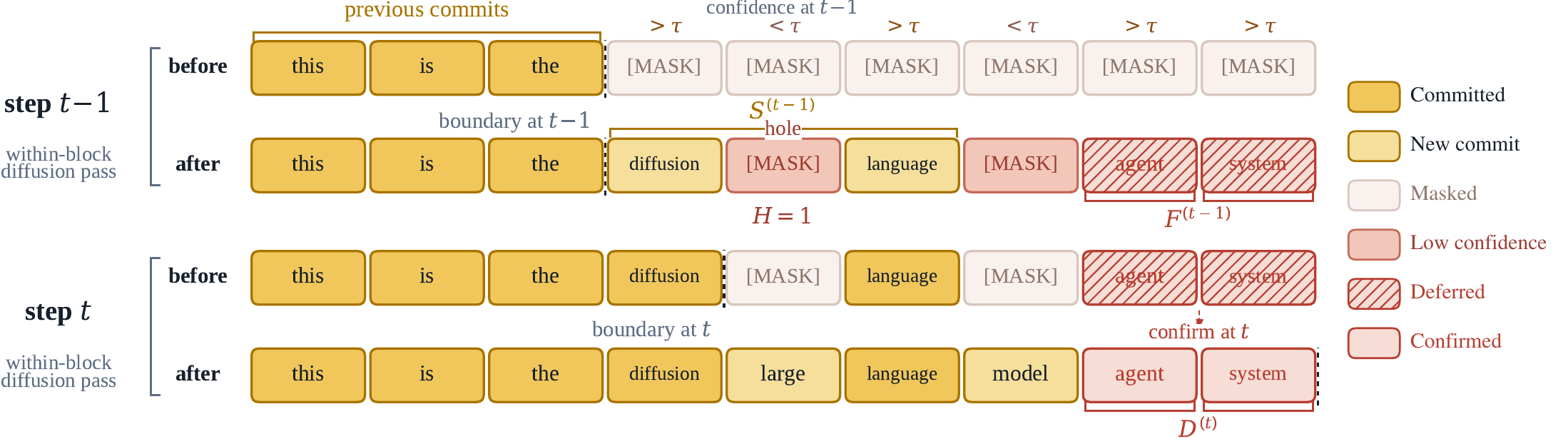}
\caption{\textbf{CCTC combines spatial structure with temporal confirmation.}
The boundary-anchored core advances through nearby high-confidence predictions, while
proposals outside the core are revisited under updated context before confirmation. This couples parallel commitment
to the evolving context: confidence identifies eligible positions, and position and persistence
guide when to commit them as the surrounding context develops over successive decoding steps.}\vspace{-2mm}
\label{fig:cctc-compare}
\end{figure*}

While CVEE decides when to end generation, the preceding blocks still need to be decoded
efficiently. CCTC addresses this remaining cost by building on Fast-dLLM's confidence-threshold
parallel commitment rule \citep{wu2025fastdllm}. Specifically, it extends this selection rule with a boundary-anchored
core and deferred confirmation: eligible positions outside the core are checked again under
updated context (Figure~\ref{fig:cctc-compare}). This spatial prior is motivated by the decoding order studied in Efficient-DLM
\citep{fu2026efficientdlm} and blockwise prediction in DiffusionVL \citep{zeng2026diffusionvl}. CCTC uses this prior to organize additional commitments within
the active block while retaining parallel prediction and the reference decoder's scheduled updates.

\noindent\textbf{Selecting eligible positions.}
To formalize this selection, let $E^{(t)}$ be the currently masked positions in active block $B_n$ whose confidence is at
least $\tau_{\text{CCTC}}$ at step $t$. CCTC partitions these positions into a core for immediate
commitment and deferred proposals for confirmation under the next step's updated context.
This separates two complementary checks: the core uses position to select immediate updates,
whereas confirmation tests whether deferred predictions persist across steps.

\noindent\textbf{Boundary-anchored commitment within a block.}
The core favors additional commitments near the resolved left boundary while predictions
remain parallel. More precisely, the gap budget $H$ specifies how many unresolved low-confidence positions
it may cross:\vspace{-1mm}
\begin{equation}
h_i^{(t)} =
\big|\{\,j\in B_n : j<i,\ x_j^{(t)}=\texttt{[MASK]},\ j\notin E^{(t)}\,\}\big|,
\qquad
S^{(t)}=\{\,i\in E^{(t)}:h_i^{(t)}\leq H\,\}.\vspace{-1mm}
\label{eq:cctc}
\end{equation}
Here $h_i^{(t)}$ counts unresolved holes before $i$. Already committed positions do not count, so
the core remains anchored at the resolved boundary. Eligible positions outside the core are \textbf{deferred} unless confirmed or selected by the
scheduled rule; their predictions are recorded for the next step. The boundary constraint, however, applies
only to the core: confirmed deferrals and scheduled commits can bypass it, so the full update
is not strictly left to right.

\noindent\textbf{Confirming proposals under updated context.}
\label{sec:cctc}
To reconsider these deferred proposals after the context changes, let $F^{(t-1)}=E^{(t-1)}\setminus\big(S^{(t-1)}\cup D^{(t-1)}\big)$ be the proposal buffer
from $t\!-\!1$, with recorded predictions $\tilde{x}_i^{(t-1)}$. Positions already committed by
the schedule are excluded by the mask condition below. At step $t$, confirmation selects\vspace{-1mm}
\begin{equation}
D^{(t)}=\{\,i\in F^{(t-1)}\cap E^{(t)} : x_i^{(t)}=\texttt{[MASK]},\ \hat{x}_i^{(t)}=\tilde{x}_i^{(t-1)}\,\},\vspace{-1mm}
\label{eq:confirm}
\end{equation}
so this branch commits a deferred position only if it still clears $\tau_{\text{CCTC}}$ and still predicts the
recorded token under the next step\textquotesingle s updated context. If a prediction changes or its confidence falls, however, the position receives no commit
from this branch. It may still commit through the core or
scheduled top-$k$ rule. If it remains above the threshold without entering the core, it is
deferred again with its new prediction, and otherwise it leaves the deferral buffer. The proposal buffer is empty at block entry and never carries across blocks. Because confirmation reuses the next scheduled forward pass, it requires no additional model
evaluation; committed tokens remain fixed. Combining these branches with the reference
schedule gives the complete update. We accelerate all blocks, $\mathcal{A}=\{0,\ldots,N{-}1\}$, and let $\mathcal{K}^{(t)}$ denote the
reference decoder's scheduled top-$k$ selection. Every masked $i\in B_n$, $n\in\mathcal{A}$,
updates as\vspace{-3mm}
\begin{equation}
x_i^{(t+1)} =
\left\{
\begin{array}{@{}l@{\;}l@{\qquad}l@{\;}l@{}}
\hat{x}_i^{(t)}, & i \in S^{(t)} \ \text{(a) core} &
\hat{x}_i^{(t)}, & i \in D^{(t)} \ \text{(b) confirmed deferral} \\[2pt]
\hat{x}_i^{(t)}, & i \in \mathcal{K}^{(t)} \ \text{(c) scheduled top-}k &
x_i^{(t)}, & \text{otherwise} \ \text{(d) stays masked}.
\end{array}
\right.\vspace{-1mm}
\label{eq:blockaccel}
\end{equation}
CCTC constrains only the additional threshold-driven commits in branches (a) and (b), while the
reference decoder's scheduled top-$k$ branch (c) is retained unchanged to guarantee progress.
Once all positions in a block have committed, the block-completion condition
$R_n(x)=\bigwedge_{i\in B_n}[x_i\neq\texttt{[MASK]}]$ holds, and the remaining
steps allocated to that block are skipped.

Together, these decisions determine both the pace and duration of decoding: CCTC shortens
the path through blocks that must be generated, while CVEE determines how
far along that path decoding needs to proceed. Committed tokens become context for later
predictions, so CVEE observes the trajectory produced by local acceleration. The resulting
decoder combines parallel block completion with revision-adaptive termination, without
additional training or extra forward passes for confirmation.

%% file: sections/experiments.tex
\vspace{-2mm}\section{Experiments}
\label{sec:experiments}
\vspace{-2mm}

\begin{table}[!t]
\centering\vspace{-4mm}
\tiny
\renewcommand{\arraystretch}{0.75}
\resizebox{\textwidth}{!}{%
\begin{tabular}{llrrrrrrrrrr}
\toprule
& & \multicolumn{5}{c}{LLaDA-8B-Instruct} & \multicolumn{5}{c}{Dream-7B-Instruct} \\
\cmidrule(lr){3-7}\cmidrule(lr){8-12}
Task & Variant & Acc (\%) & Avg.\ Step & Saving & TPS & Speedup & Acc (\%) & Avg.\ Step & Saving & TPS & Speedup \\
\midrule
\multicolumn{12}{l}{\textit{General / short-answer tasks (single-token or short-span answers)}} \\
\midrule
MMLU & Baseline & 64.0 & 64.0 & $0.0\%$ & 30.4 & 1.00$\times$ & 71.5 & 64.0 & $0.0\%$ & 28.2 & 1.00$\times$ \\
(n=200, sequence length=64) & Prophet \citep{li2026diffusion} & 62.5 (-1.5) & 8.2 (-55.8) & $87.1\%$ & 202.2 & 6.65$\times$ & 71.5 (+0.0) & 14.9 (-49.2) & $76.8\%$ & 109.8 & 3.89$\times$ \\
(block=16, 4 blocks) & SchED \citep{mohamed2025sched} & \textcolor{failred}{61.5} (-2.5) & 31.0 (-33.0) & $51.5\%$ & 53.3 & 1.75$\times$ & \textcolor{failred}{69.0} (-2.5) & 13.0 (-51.1) & $79.8\%$ & 178.7 & 6.34$\times$ \\
 & SlowFast \citep{wei2025slowfast} & 63.0 (-1.0) & 17.9 (-46.1) & $72.0\%$ & 130.4 & 4.29$\times$ & 74.5 (+3.0) & 9.6 (-54.4) & $85.0\%$ & 196.9 & 6.98$\times$ \\
 & KLASS \citep{kim2025klass} & 62.5 (-1.5) & 29.2 (-34.8) & $54.4\%$ & 51.7 & 1.70$\times$ & 72.5 (+1.0) & 11.3 (-52.7) & $82.3\%$ & 129.0 & 4.57$\times$ \\
 & WINO \citep{hong2026wino} & \textcolor{failred}{61.0} (-3.0) & 16.8 (-47.2) & $73.8\%$ & 100.4 & 3.30$\times$ & -- & -- & -- & -- & -- \\
\rowcolor{finalres}
 & \textcolor{oursemph}{$\mathrm{C}^4$} & 64.0 (+0.0) & 4.8 (-59.2) & $92.5\%$ & 490.4 & 16.13$\times$ & 70.0 (-1.5) & 4.5 (-59.5) & $93.0\%$ & 389.7 & 13.82$\times$ \\
\midrule
ARC-C & Baseline & 86.5 & 64.0 & $0.0\%$ & 33.8 & 1.00$\times$ & 87.5 & 64.0 & $0.0\%$ & 32.6 & 1.00$\times$ \\
(n=200, sequence length=64) & Prophet \citep{li2026diffusion} & 86.0 (-0.5) & 7.5 (-56.5) & $88.3\%$ & 295.3 & 8.74$\times$ & 86.0 (-1.5) & 14.0 (-50.0) & $78.2\%$ & 138.3 & 4.24$\times$ \\
(block=16, 4 blocks) & SchED \citep{mohamed2025sched} & 85.0 (-1.5) & 41.1 (-22.9) & $35.8\%$ & 57.5 & 1.70$\times$ & 88.5 (+1.0) & 11.0 (-53.0) & $82.8\%$ & 215.1 & 6.60$\times$ \\
 & SlowFast \citep{wei2025slowfast} & 85.0 (-1.5) & 23.7 (-40.3) & $63.0\%$ & 107.3 & 3.17$\times$ & 87.0 (-0.5) & 10.4 (-53.6) & $83.8\%$ & 219.4 & 6.73$\times$ \\
 & KLASS \citep{kim2025klass} & 87.0 (+0.5) & 37.5 (-26.5) & $41.5\%$ & 53.7 & 1.59$\times$ & 88.5 (+1.0) & 11.6 (-52.4) & $81.9\%$ & 158.9 & 4.87$\times$ \\
 & WINO \citep{hong2026wino} & \textcolor{failred}{84.0} (-2.5) & 20.1 (-43.9) & $68.6\%$ & 76.9 & 2.28$\times$ & -- & -- & -- & -- & -- \\
\rowcolor{finalres}
 & \textcolor{oursemph}{$\mathrm{C}^4$} & 86.5 (+0.0) & 3.9 (-60.1) & $94.0\%$ & 483.3 & 14.30$\times$ & 88.5 (+1.0) & 4.2 (-59.8) & $93.5\%$ & 565.7 & 17.35$\times$ \\
\midrule
HellaSwag & Baseline & 75.5 & 64.0 & $0.0\%$ & 20.8 & 1.00$\times$ & 75.5 & 64.0 & $0.0\%$ & 22.1 & 1.00$\times$ \\
(n=200, sequence length=64) & Prophet \citep{li2026diffusion} & 75.0 (-0.5) & 6.0 (-58.0) & $90.6\%$ & 223.4 & 10.74$\times$ & 75.0 (-0.5) & 16.2 (-47.8) & $74.7\%$ & 82.0 & 3.71$\times$ \\
(block=16, 4 blocks) & SchED \citep{mohamed2025sched} & 75.5 (+0.0) & 27.7 (-36.3) & $56.7\%$ & 46.4 & 2.23$\times$ & 75.0 (-0.5) & 19.0 (-45.0) & $70.3\%$ & 63.0 & 2.85$\times$ \\
 & SlowFast \citep{wei2025slowfast} & 77.0 (+1.5) & 10.9 (-53.1) & $83.0\%$ & 134.2 & 6.45$\times$ & \textcolor{failred}{72.5} (-3.0) & 9.0 (-55.0) & $85.9\%$ & 160.0 & 7.24$\times$ \\
 & KLASS \citep{kim2025klass} & 75.0 (-0.5) & 19.4 (-44.6) & $69.6\%$ & 56.5 & 2.72$\times$ & \textcolor{failred}{69.5} (-6.0) & 11.3 (-52.7) & $82.3\%$ & 94.0 & 4.25$\times$ \\
 & WINO \citep{hong2026wino} & \textcolor{failred}{73.0} (-2.5) & 12.9 (-51.1) & $79.8\%$ & 79.0 & 3.80$\times$ & -- & -- & -- & -- & -- \\
\rowcolor{finalres}
 & \textcolor{oursemph}{$\mathrm{C}^4$} & 75.5 (+0.0) & 3.7 (-60.3) & $94.3\%$ & 348.6 & 16.76$\times$ & 75.5 (+0.0) & 4.4 (-59.6) & $93.1\%$ & 297.8 & 13.48$\times$ \\
\midrule
WinoGrande & Baseline & 75.5 & 64.0 & $0.0\%$ & 36.1 & 1.00$\times$ & 71.5 & 64.0 & $0.0\%$ & 35.5 & 1.00$\times$ \\
(n=200, sequence length=64) & Prophet \citep{li2026diffusion} & 76.0 (+0.5) & 5.9 (-58.1) & $90.8\%$ & 383.9 & 10.63$\times$ & 69.5 (-2.0) & 15.3 (-48.7) & $76.1\%$ & 140.6 & 3.96$\times$ \\
(block=16, 4 blocks) & SchED \citep{mohamed2025sched} & 75.5 (+0.0) & 5.8 (-58.2) & $91.0\%$ & 447.1 & 12.39$\times$ & 71.0 (-0.5) & 2.1 (-61.9) & $96.7\%$ & 1034.4 & 29.14$\times$ \\
 & SlowFast \citep{wei2025slowfast} & 77.5 (+2.0) & 8.5 (-55.5) & $86.7\%$ & 285.5 & 7.91$\times$ & 71.0 (-0.5) & 8.0 (-56.0) & $87.5\%$ & 279.6 & 7.88$\times$ \\
 & KLASS \citep{kim2025klass} & 76.5 (+1.0) & 13.3 (-50.7) & $79.2\%$ & 160.0 & 4.43$\times$ & 72.5 (+1.0) & 7.6 (-56.4) & $88.2\%$ & 236.8 & 6.67$\times$ \\
 & WINO \citep{hong2026wino} & 77.0 (+1.5) & 10.1 (-53.9) & $84.2\%$ & 176.1 & 4.88$\times$ & -- & -- & -- & -- & -- \\
\rowcolor{finalres}
 & \textcolor{oursemph}{$\mathrm{C}^4$} & 75.5 (+0.0) & 3.4 (-60.6) & $94.6\%$ & 672.8 & 18.64$\times$ & 71.5 (+0.0) & 4.2 (-59.8) & $93.4\%$ & 561.4 & 15.81$\times$ \\
\midrule
PIQA & Baseline & 81.5 & 64.0 & $0.0\%$ & 35.2 & 1.00$\times$ & 87.5 & 64.0 & $0.0\%$ & 32.5 & 1.00$\times$ \\
(n=200, sequence length=64) & Prophet \citep{li2026diffusion} & 82.0 (+0.5) & 10.8 (-53.2) & $83.1\%$ & 245.7 & 6.98$\times$ & 87.5 (+0.0) & 18.9 (-45.2) & $70.5\%$ & 104.0 & 3.20$\times$ \\
(block=16, 4 blocks) & SchED \citep{mohamed2025sched} & 81.0 (-0.5) & 24.5 (-39.5) & $61.7\%$ & 98.6 & 2.80$\times$ & 87.0 (-0.5) & 17.1 (-46.9) & $73.3\%$ & 129.1 & 3.97$\times$ \\
 & SlowFast \citep{wei2025slowfast} & 84.0 (+2.5) & 10.0 (-54.0) & $84.4\%$ & 205.2 & 5.83$\times$ & 89.5 (+2.0) & 9.1 (-55.0) & $85.9\%$ & 209.2 & 6.44$\times$ \\
 & KLASS \citep{kim2025klass} & 81.5 (+0.0) & 18.6 (-45.4) & $70.9\%$ & 106.3 & 3.02$\times$ & 91.5 (+4.0) & 10.6 (-53.4) & $83.4\%$ & 158.0 & 4.86$\times$ \\
 & WINO \citep{hong2026wino} & 81.0 (-0.5) & 11.4 (-52.6) & $82.2\%$ & 187.7 & 5.33$\times$ & -- & -- & -- & -- & -- \\
\rowcolor{finalres}
 & \textcolor{oursemph}{$\mathrm{C}^4$} & 82.0 (+0.5) & 3.6 (-60.4) & $94.4\%$ & 546.5 & 15.53$\times$ & 88.0 (+0.5) & 3.5 (-60.5) & $94.5\%$ & 575.7 & 17.71$\times$ \\
\midrule
\multicolumn{12}{l}{\textit{Long-reasoning tasks (multi-step CoT)}} \\
\midrule
GSM8K & Baseline & 66.0 & 256.0 & $0.0\%$ & 19.0 & 1.00$\times$ & 85.0 & 256.0 & $0.0\%$ & 21.2 & 1.00$\times$ \\
(n=100, sequence length=256) & Prophet \citep{li2026diffusion} & \textcolor{failred}{38.0} (-28.0) & 185.0 (-71.0) & $27.7\%$ & 22.6 & 1.19$\times$ & \textcolor{failred}{16.0} (-69.0) & 76.1 (-179.9) & $70.3\%$ & 67.0 & 3.16$\times$ \\
(block=32, 8 blocks) & SchED \citep{mohamed2025sched} & \textcolor{failred}{59.0} (-7.0) & 224.3 (-31.7) & $12.4\%$ & 21.1 & 1.11$\times$ & \textcolor{failred}{18.0} (-67.0) & 76.9 (-179.1) & $70.0\%$ & 64.9 & 3.06$\times$ \\
 & SlowFast \citep{wei2025slowfast} & \textcolor{failred}{41.0} (-25.0) & 70.9 (-185.1) & $72.3\%$ & 68.0 & 3.58$\times$ & \textcolor{failred}{20.0} (-65.0) & 89.2 (-166.8) & $65.1\%$ & 59.6 & 2.81$\times$ \\
 & KLASS \citep{kim2025klass} & 65.0 (-1.0) & 105.5 (-150.5) & $58.8\%$ & 34.5 & 1.82$\times$ & \textcolor{failred}{33.0} (-52.0) & 96.5 (-159.5) & $62.3\%$ & 37.2 & 1.75$\times$ \\
 & WINO \citep{hong2026wino} & 72.0 (+6.0) & 49.2 (-206.8) & $80.8\%$ & 76.6 & 4.03$\times$ & -- & -- & -- & -- & -- \\
\rowcolor{finalres}
 & \textcolor{oursemph}{$\mathrm{C}^4$} & 64.0 (-2.0) & 62.5 (-193.5) & $75.6\%$ & 67.5 & 3.55$\times$ & 85.0 (+0.0) & 54.5 (-201.5) & $78.7\%$ & 102.1 & 4.82$\times$ \\
\midrule
MATH & Baseline & 32.0 & 256.0 & $0.0\%$ & 17.8 & 1.00$\times$ & 42.0 & 256.0 & $0.0\%$ & 19.4 & 1.00$\times$ \\
(n=100, sequence length=256) & Prophet \citep{li2026diffusion} & \textcolor{failred}{19.0} (-13.0) & 192.7 (-63.3) & $24.7\%$ & 21.4 & 1.20$\times$ & \textcolor{failred}{15.0} (-27.0) & 164.9 (-91.1) & $35.6\%$ & 26.7 & 1.38$\times$ \\
(block=32, 8 blocks) & SchED \citep{mohamed2025sched} & \textcolor{failred}{26.0} (-6.0) & 233.5 (-22.5) & $8.8\%$ & 19.3 & 1.08$\times$ & \textcolor{failred}{33.0} (-9.0) & 199.0 (-57.0) & $22.3\%$ & 23.3 & 1.20$\times$ \\
 & SlowFast \citep{wei2025slowfast} & \textcolor{failred}{24.0} (-8.0) & 100.0 (-156.0) & $60.9\%$ & 45.5 & 2.56$\times$ & \textcolor{failred}{9.0} (-33.0) & 98.9 (-157.1) & $61.4\%$ & 51.9 & 2.68$\times$ \\
 & KLASS \citep{kim2025klass} & 34.0 (+2.0) & 133.9 (-122.1) & $47.7\%$ & 26.1 & 1.47$\times$ & \textcolor{failred}{4.0} (-38.0) & 98.7 (-157.3) & $61.4\%$ & 40.3 & 2.08$\times$ \\
 & WINO \citep{hong2026wino} & 35.0 (+3.0) & 81.7 (-174.3) & $68.1\%$ & 42.8 & 2.40$\times$ & -- & -- & -- & -- & -- \\
\rowcolor{finalres}
 & \textcolor{oursemph}{$\mathrm{C}^4$} & 33.0 (+1.0) & 92.5 (-163.5) & $63.9\%$ & 46.7 & 2.62$\times$ & 42.0 (+0.0) & 84.6 (-171.5) & $67.0\%$ & 53.5 & 2.76$\times$ \\
\midrule
SVAMP & Baseline & 81.0 & 256.0 & $0.0\%$ & 19.3 & 1.00$\times$ & 76.0 & 256.0 & $0.0\%$ & 21.4 & 1.00$\times$ \\
(n=100, sequence length=256) & Prophet \citep{li2026diffusion} & \textcolor{failred}{40.0} (-41.0) & 160.1 (-96.0) & $37.5\%$ & 27.5 & 1.42$\times$ & \textcolor{failred}{29.0} (-47.0) & 48.8 (-207.2) & $80.9\%$ & 76.1 & 3.56$\times$ \\
(block=32, 8 blocks) & SchED \citep{mohamed2025sched} & \textcolor{failred}{70.0} (-11.0) & 196.0 (-60.0) & $23.4\%$ & 26.9 & 1.39$\times$ & \textcolor{failred}{20.0} (-56.0) & 42.9 (-213.1) & $83.2\%$ & 75.8 & 3.54$\times$ \\
 & SlowFast \citep{wei2025slowfast} & \textcolor{failred}{46.0} (-35.0) & 81.7 (-174.3) & $68.1\%$ & 63.4 & 3.28$\times$ & \textcolor{failred}{67.0} (-9.0) & 49.9 (-206.1) & $80.5\%$ & 137.0 & 6.40$\times$ \\
 & KLASS \citep{kim2025klass} & 79.0 (-2.0) & 96.4 (-159.6) & $62.4\%$ & 39.9 & 2.07$\times$ & \textcolor{failred}{71.0} (-5.0) & 58.1 (-197.9) & $77.3\%$ & 71.0 & 3.32$\times$ \\
 & WINO \citep{hong2026wino} & 85.0 (+4.0) & 45.6 (-210.4) & $82.2\%$ & 81.0 & 4.20$\times$ & -- & -- & -- & -- & -- \\
\rowcolor{finalres}
 & \textcolor{oursemph}{$\mathrm{C}^4$} & 81.0 (+0.0) & 58.8 (-197.2) & $77.0\%$ & 82.8 & 4.29$\times$ & 75.0 (-1.0) & 42.8 (-213.2) & $83.3\%$ & 113.0 & 5.28$\times$ \\
\midrule
ASDiv & Baseline & 68.0 & 256.0 & $0.0\%$ & 19.4 & 1.00$\times$ & 78.0 & 256.0 & $0.0\%$ & 21.4 & 1.00$\times$ \\
(n=100, sequence length=256) & Prophet \citep{li2026diffusion} & \textcolor{failred}{37.0} (-31.0) & 163.7 (-92.3) & $36.1\%$ & 29.5 & 1.52$\times$ & \textcolor{failred}{25.0} (-53.0) & 54.8 (-201.3) & $78.6\%$ & 122.5 & 5.72$\times$ \\
(block=32, 8 blocks) & SchED \citep{mohamed2025sched} & \textcolor{failred}{60.0} (-8.0) & 197.0 (-59.0) & $23.0\%$ & 28.6 & 1.47$\times$ & \textcolor{failred}{17.0} (-61.0) & 53.3 (-202.7) & $79.2\%$ & 101.1 & 4.72$\times$ \\
 & SlowFast \citep{wei2025slowfast} & \textcolor{failred}{34.0} (-34.0) & 82.4 (-173.7) & $67.8\%$ & 64.9 & 3.35$\times$ & \textcolor{failred}{52.0} (-26.0) & 57.9 (-198.1) & $77.4\%$ & 144.0 & 6.73$\times$ \\
 & KLASS \citep{kim2025klass} & 66.0 (-2.0) & 100.8 (-155.2) & $60.6\%$ & 40.7 & 2.10$\times$ & \textcolor{failred}{58.0} (-20.0) & 67.2 (-188.8) & $73.7\%$ & 65.4 & 3.06$\times$ \\
 & WINO \citep{hong2026wino} & \textcolor{failred}{60.0} (-8.0) & 48.2 (-207.8) & $81.2\%$ & 84.5 & 4.36$\times$ & -- & -- & -- & -- & -- \\
\rowcolor{finalres}
 & \textcolor{oursemph}{$\mathrm{C}^4$} & 71.0 (+3.0) & 60.9 (-195.1) & $76.2\%$ & 86.9 & 4.48$\times$ & 76.0 (-2.0) & 45.8 (-210.2) & $82.1\%$ & 135.1 & 6.31$\times$ \\
\midrule
GSM-Hard & Baseline & 35.0 & 256.0 & $0.0\%$ & 18.8 & 1.00$\times$ & 42.0 & 256.0 & $0.0\%$ & 20.6 & 1.00$\times$ \\
(n=100, sequence length=256) & Prophet \citep{li2026diffusion} & \textcolor{failred}{18.0} (-17.0) & 194.1 (-61.9) & $24.2\%$ & 21.9 & 1.16$\times$ & \textcolor{failred}{9.0} (-33.0) & 103.9 (-152.1) & $59.4\%$ & 45.7 & 2.22$\times$ \\
(block=32, 8 blocks) & SchED \citep{mohamed2025sched} & \textcolor{failred}{31.0} (-4.0) & 237.4 (-18.6) & $7.3\%$ & 19.7 & 1.05$\times$ & \textcolor{failred}{10.0} (-32.0) & 123.5 (-132.6) & $51.8\%$ & 41.6 & 2.02$\times$ \\
 & SlowFast \citep{wei2025slowfast} & \textcolor{failred}{21.0} (-14.0) & 70.0 (-186.0) & $72.6\%$ & 70.2 & 3.73$\times$ & \textcolor{failred}{11.0} (-31.0) & 96.8 (-159.2) & $62.2\%$ & 62.6 & 3.04$\times$ \\
 & KLASS \citep{kim2025klass} & 35.0 (+0.0) & 109.7 (-146.3) & $57.1\%$ & 34.0 & 1.81$\times$ & \textcolor{failred}{10.0} (-32.0) & 111.7 (-144.3) & $56.4\%$ & 35.1 & 1.70$\times$ \\
 & WINO \citep{hong2026wino} & 34.0 (-1.0) & 51.2 (-204.8) & $80.0\%$ & 77.3 & 4.11$\times$ & -- & -- & -- & -- & -- \\
\rowcolor{finalres}
 & \textcolor{oursemph}{$\mathrm{C}^4$} & 37.0 (+2.0) & 64.4 (-191.6) & $74.8\%$ & 67.6 & 3.60$\times$ & 45.0 (+3.0) & 64.9 (-191.1) & $74.7\%$ & 82.2 & 3.99$\times$ \\
\midrule
\multicolumn{12}{l}{\textit{Code generation (whole-program answer, execution-graded)}} \\
\midrule
HumanEval & Baseline & 40.2 & 384.0 & $0.0\%$ & 13.8 & 1.00$\times$ & 45.1 & 384.0 & $0.0\%$ & 14.5 & 1.00$\times$ \\
(n=164, sequence length=384) & Prophet \citep{li2026diffusion} & \textcolor{failred}{3.0} (-37.2) & 124.0 (-260.0) & $67.7\%$ & 30.9 & 2.24$\times$ & \textcolor{failred}{0.6} (-44.5) & 29.3 (-354.7) & $92.4\%$ & 144.2 & 9.94$\times$ \\
(block=32, 12 blocks) & SchED \citep{mohamed2025sched} & \textcolor{failred}{12.8} (-27.4) & 175.7 (-208.3) & $54.3\%$ & 33.1 & 2.40$\times$ & \textcolor{failred}{1.8} (-43.3) & 39.3 (-344.7) & $89.8\%$ & 142.9 & 9.86$\times$ \\
 & SlowFast \citep{wei2025slowfast} & \textcolor{failred}{19.5} (-20.7) & 36.3 (-347.7) & $90.5\%$ & 213.1 & 15.44$\times$ & \textcolor{failred}{23.2} (-22.0) & 36.4 (-347.6) & $90.5\%$ & 205.6 & 14.18$\times$ \\
 & KLASS \citep{kim2025klass} & 40.9 (+0.6) & 92.6 (-291.4) & $75.9\%$ & 45.4 & 3.29$\times$ & \textcolor{failred}{25.6} (-19.5) & 103.5 (-280.5) & $73.1\%$ & 34.7 & 2.39$\times$ \\
 & WINO \citep{hong2026wino} & \textcolor{failred}{37.8} (-2.4) & 54.2 (-329.8) & $85.9\%$ & 77.0 & 5.58$\times$ & -- & -- & -- & -- & -- \\
\rowcolor{finalres}
 & \textcolor{oursemph}{$\mathrm{C}^4$} & 42.1 (+1.8) & 39.4 (-344.6) & $89.7\%$ & 120.0 & 8.70$\times$ & 43.3 (-1.8) & 29.2 (-354.8) & $92.4\%$ & 146.0 & 10.07$\times$ \\
\midrule
MBPP & Baseline & 45.9 & 384.0 & $0.0\%$ & 15.4 & 1.00$\times$ & 56.0 & 384.0 & $0.0\%$ & 15.6 & 1.00$\times$ \\
(n=257, sequence length=384) & Prophet \citep{li2026diffusion} & \textcolor{failred}{1.9} (-44.0) & 32.1 (-351.9) & $91.6\%$ & 73.2 & 4.75$\times$ & \textcolor{failred}{0.8} (-55.3) & 1.4 (-382.6) & $99.6\%$ & 433.7 & 27.80$\times$ \\
(block=32, 12 blocks) & SchED \citep{mohamed2025sched} & \textcolor{failred}{27.6} (-18.3) & 66.6 (-317.4) & $82.6\%$ & 101.3 & 6.58$\times$ & \textcolor{failred}{1.9} (-54.1) & 16.3 (-367.7) & $95.8\%$ & 414.1 & 26.54$\times$ \\
 & SlowFast \citep{wei2025slowfast} & \textcolor{failred}{41.2} (-4.7) & 21.6 (-362.4) & $94.4\%$ & 362.0 & 23.51$\times$ & \textcolor{failred}{46.7} (-9.3) & 30.0 (-354.0) & $92.2\%$ & 274.1 & 17.57$\times$ \\
 & KLASS \citep{kim2025klass} & 46.5 (+0.6) & 69.2 (-314.8) & $82.0\%$ & 66.8 & 4.34$\times$ & \textcolor{failred}{53.0} (-3.0) & 93.1 (-290.9) & $75.8\%$ & 42.3 & 2.71$\times$ \\
 & WINO \citep{hong2026wino} & 47.1 (+1.2) & 54.4 (-329.6) & $85.8\%$ & 88.0 & 5.71$\times$ & -- & -- & -- & -- & -- \\
\rowcolor{finalres}
 & \textcolor{oursemph}{$\mathrm{C}^4$} & 45.9 (+0.0) & 33.3 (-350.7) & $91.3\%$ & 170.8 & 11.09$\times$ & 56.0 (+0.0) & 25.5 (-358.5) & $93.4\%$ & 209.3 & 13.42$\times$ \\
\bottomrule
\end{tabular}%
}
\caption{\textbf{Zero-shot evaluation across 12 tasks on LLaDA-8B-Instruct and Dream-7B-Instruct.}
Parentheses report accuracy and step changes from full decoding. TPS denotes tokens per second;
speedup is the method/baseline TPS ratio. \textcolor{failred}{Red} marks accuracy drops exceeding $2.0$ percentage points.}
\label{tab:main}\vspace{-3mm}
\end{table}

\label{sec:setup}
\vspace{-1mm}
\noindent\textbf{Benchmarks and hardware.}
We evaluate LLaDA-8B-Instruct \citep{nie2025llada} and Dream-7B-Instruct
\citep{ye2025dream} with bf16 and greedy decoding; main-table timing uses a single RTX~4090.
The $12$ tasks comprise five \emph{short-answer} benchmarks (MMLU
\citep{hendrycks2021mmlu}, ARC-Challenge \citep{clark2018arc}, HellaSwag
\citep{zellers2019hellaswag}, WinoGrande \citep{sakaguchi2021winogrande}, and PIQA
\citep{bisk2020piqa}), five \emph{long-reasoning} benchmarks (GSM8K
\citep{cobbe2021gsm8k}, MATH \citep{hendrycks2021math}, SVAMP
\citep{patel2021svamp}, ASDiv \citep{miao2020asdiv}, and GSM-Hard
\citep{gao2023pal}), and two \emph{code-generation} benchmarks graded by hidden tests (HumanEval
\citep{chen2021humaneval} and MBPP \citep{austin2021mbpp}).

\vspace{-1mm}
\noindent\textbf{Hyperparameters and calibration.}
$\mathrm{C}^4$ uses five gate and commitment hyperparameters in the main comparison.
CVEE uses $\gamma{=}2.0$, $\tau_{\mathrm{CVEE}}{=}0.7$, and $p_{\min}{=}3$;
CCTC uses $\tau_{\mathrm{CCTC}}{=}0.8$ and gap budget $H{=}1$.
The three CVEE parameters are calibrated once on a shared $120$-example LLaDA pool and then
frozen across all $12$ tasks and transferred unchanged to Dream. The two CCTC parameters are fixed
\emph{a priori}. All reported configurations are evaluated with live decoding.
Appendix~\ref{app:hyperparams} details the prompting protocol, baseline configurations, and task-specific extraction rules.

\noindent\textbf{Evaluation metrics.}
Accuracy follows task-specific answer extraction, or execution grading for code.
Step is the number of executed decoding steps, and
Saving is $1-\text{Step}/T_{\mathrm{full}}$.
Speedup is method TPS divided by the corresponding full-decoding baseline TPS.
Table~\ref{tab:main} highlights drops exceeding $2.0$ points as a visual aid, independent of
calibration. Score differences are finite-sample point estimates, not equivalence claims.
Accuracy and step counts use the task-specific evaluation sets; throughput uses a fixed $20$-example timing subset with three repeats (Appendix~\ref{app:retiming}).

\vspace{-2mm}
\subsection{Performance across Task Regimes}
\label{sec:generalization}\vspace{-1mm}

\begin{figure}[!ht]
\centering\vspace{-4mm}
\includegraphics[width=\textwidth]{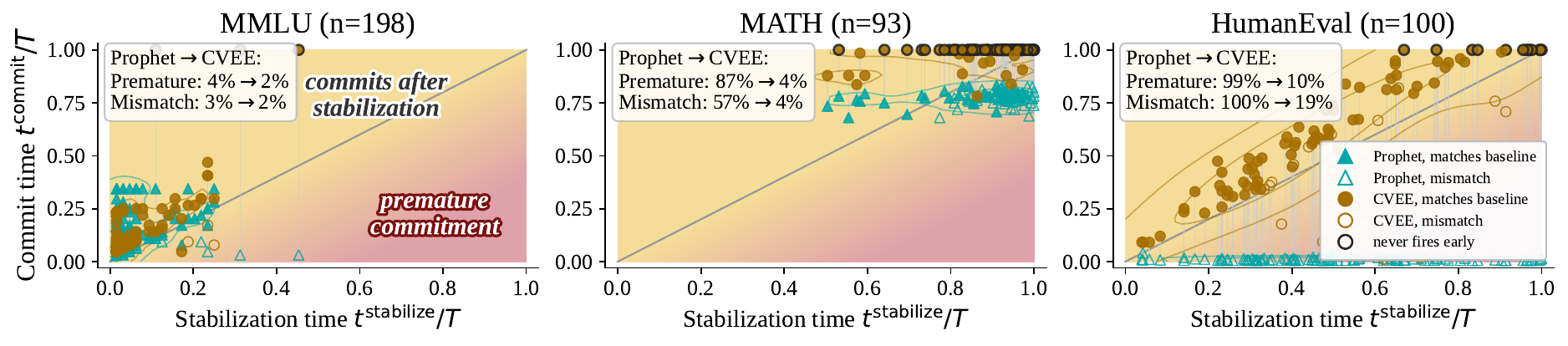}
\caption{\textbf{Exit--stabilization phase diagram.}
Each faint line pairs Prophet and CVEE on the same LLaDA-8B-Instruct trajectory.
On MMLU, both often trigger near answer stabilization; on MATH and HumanEval, Prophet frequently
triggers before the final answer has stabilized, whereas CVEE waits substantially longer.
The comparison measures alignment between stopping and answer stabilization.}
\label{fig:phase-diagram}\vspace{-3mm}
\end{figure}

Across the $12$ tasks in Table~\ref{tab:main}, $\mathrm{C}^4$ reduces decoding cost on both
backbones while largely preserving task performance. To understand where the acceleration
arises, we relate these savings to the timing of stabilization. The dominant source changes
across tasks: early-settling predictions leave later blocks to skip, whereas
extended generation requires faster completion of the blocks that remain necessary.
Figure~\ref{fig:phase-diagram} relates stopping to stabilization on individual trajectories;
the component ablations in Section~\ref{sec:blockaccel-ablation} separate the two sources of savings.

\vspace{-1mm}
\noindent\textbf{Early stabilization favors global stopping.}
On short-answer tasks, predictions often settle within the first block. CVEE can then skip
subsequent blocks, making global termination the main source of savings. The MMLU trajectories
in Figure~\ref{fig:phase-diagram} place stopping close to stabilization, illustrating how a short
useful decoding trajectory can coexist with a much larger prescribed generation budget.

\vspace{-1mm}
\noindent\textbf{Continued generation favors local acceleration.}
For long-reasoning tasks, however, predictions often stabilize near the decoding horizon, leaving fewer
steps for global stopping to remove. CCTC instead reduces the work inside blocks that still
need to execute. The Dream GSM8K ablation makes this distinction clear: CVEE alone saves few
steps, while adding CCTC removes most of the within-block work. Consequently, local sampling
quality becomes especially important in this regime. For example, WINO outperforms $\mathrm{C}^4$ on LLaDA GSM8K in both
accuracy ($72.0\%$ versus $64.0\%$) and speedup ($4.03\times$ versus $3.55\times$), showing
that the strongest local sampler can retain an advantage when early-exit opportunities are scarce.

\vspace{-1mm}
\noindent\textbf{Code generation links local updates to program completion.}
Code generation adds a further dependency: correctness is determined by execution of the
complete program, so commitments throughout
the output contribute to the evaluated result. The HumanEval ablations show that accelerating
the final block provides further savings. At the same time, its later stabilization in
Figure~\ref{fig:phase-diagram} highlights the need to distinguish faster token commitment from
readiness to terminate: CVEE waits longer than Prophet while the program continues evolving.

\vspace{-1mm}
Beyond the overall savings, we examine how commitment settings and locations affect this
balance. Threshold sweeps
show a broad region of stable accuracy on short-answer tasks, while reasoning and code are
more sensitive to aggressive commitment (Appendix~\ref{app:taublk}). Paired controls further
show that protecting the monitored span changes exit timing only modestly on MMLU and GSM8K,
with identical graded answers on both backbones; protecting the entire block causes larger
changes (Appendix~\ref{app:span-control}). Together, the results connect the benefits of global
stopping to the trajectory produced by local updates, motivating the sampler-composition
experiments that follow.

\vspace{-2mm}
\subsection{Early Exit across Sampling Trajectories}
\label{sec:composition}\vspace{-1mm}

\begin{wraptable}{r}{0.52\textwidth}
\centering
\scriptsize
\renewcommand{\arraystretch}{1}
\vspace{-3mm}
\caption{\textbf{CVEE combined with WINO.}
CVEE provides substantial additional step reduction on short-answer tasks, whereas on
long-reasoning and code tasks, preserving accuracy leaves fewer opportunities to terminate the already shortened trajectory.}
\label{tab:composition}

\begin{tabular}{lrrrrrr}
\toprule
& \multicolumn{2}{c}{WINO}
& \multicolumn{2}{c}{+ CVEE}
& \multicolumn{2}{c}{Conservative CVEE} \\
Task & Acc & Step & Acc & Step & Acc & Step \\
\midrule
MMLU      & 61.0 & 16.8 & 61.5 & 4.1  & --   & --   \\
HellaSwag & 73.0 & 12.9 & 73.0 & 3.9  & --   & --   \\
GSM8K     & 72.0 & 49.2 & 36.0 & 43.6 & 70.0 & 48.5 \\
SVAMP     & 85.0 & 45.6 & 62.0 & 41.7 & 83.0 & 45.1 \\
HumanEval & 37.8 & 54.2 & 14.6 & 29.7 & 36.6 & 50.1 \\
\bottomrule
\end{tabular}

\vspace{-3mm}
\end{wraptable}

The preceding results link useful early exit to the trajectory produced by local updates.
To examine this connection with a different sampler, we pair CVEE with WINO~\citep{hong2026wino}, retaining WINO's
local update rule and using CVEE only for sequence termination (Table~\ref{tab:composition}).
This separates the stopping rule from CCTC and tests it on a different sampling trajectory.

\vspace{-1mm}
\noindent\textbf{When local acceleration leaves room to stop.}
On MMLU and HellaSwag, CVEE removes a further $70$--$76\%$ of WINO's steps with nearly
unchanged accuracy. Local sampling has shortened generation, yet predictions still settle
before the remaining scheduled updates finish. Global stopping captures this residual saving.

\vspace{-1mm}
\noindent\textbf{When the remaining steps still matter.}
On GSM8K, SVAMP, and HumanEval, however, the same exit setting substantially lowers accuracy.
A more conservative setting recovers most of the score, but brings computation close to
WINO's original step count. Thus, the apparent opportunity to terminate sooner largely
vanishes when task performance is recovered. These trajectories leave less redundant work
for CVEE to remove after local acceleration.

\vspace{-1mm}
Taken together, these composition results show that savings from sampling and stopping are
not simply additive.
A stopping rule observes predictions shaped by the local update mechanism, so its useful
operating point depends on that trajectory. Appendix~\ref{app:third-party-composition}
examines the corresponding confidence and agreement signals. We next study how CVEE and
CCTC contribute within $\mathrm{C}^4$ and how their savings change with the decoding budget.

\vspace{-3mm}
\subsection{Component Contributions and Scaling}
\label{sec:efficiency-report}
\label{sec:blockaccel-ablation}\vspace{-2mm}

Having examined how CVEE behaves on a different sampling trajectory, we now separate the
contributions of the stopping signals and commitment branches within $\mathrm{C}^4$. We then
consider how these savings change with generation length, block length, and caching. Table~\ref{tab:blockaccel-ablation} separates
the component contributions; Appendix~\ref{app:scaling-cache} examines how generation budgets,
block lengths, and caching affect the resulting accuracy--efficiency trade-offs.

\begin{table}[!t]
\centering\vspace{-4mm}
\small
\renewcommand{\arraystretch}{0.62}
\resizebox{\textwidth}{!}{%
\begin{tabular}{llrrrrrrrrrr}
\toprule
& & \multicolumn{5}{c}{LLaDA-8B-Instruct} & \multicolumn{5}{c}{Dream-7B-Instruct} \\
\cmidrule(lr){3-7}\cmidrule(lr){8-12}
Task & Variant & Acc (\%) & Step & Saving & TPS & Speedup & Acc (\%) & Step & Saving & TPS & Speedup \\
\midrule
\multicolumn{12}{l}{\textit{General / short-answer tasks}} \\
\midrule
MMLU     & Baseline & 64.0 & 64.0 & $0.0\%$ & 30.4 & 1.00$\times$ & 71.5 & 64.0 & $0.0\%$ & 28.2 & 1.00$\times$ \\
(n=200, sequence length=64)& + confidence gate only & 64.0 & 3.7 & $94.1\%$ & 636.8 & 20.95$\times$ & 70.5 & 2.9 & $95.5\%$ & 277.2 & 9.83$\times$  \\
(block=16, 4 blocks)& + stability gate only & 62.0 & 4.0 & $93.7\%$ & 525.7 & 17.29$\times$ & 70.0 & 3.7 & $94.2\%$ & 417.0 & 14.79$\times$  \\
 & + CVEE (joint gate) & 64.0 & 7.0 & $89.1\%$ & 452.7 & 14.89$\times$ & 69.5 & 8.0 & $87.5\%$ & 208.9 & 7.41$\times$ \\
 & \textcolor{oursemph}{+ CCTC, core only} & 63.5 & 5.2 & $91.9\%$ & 496.1 & 16.32$\times$ & 70.5 & 5.2 & $91.9\%$ & 323.3 & 11.46$\times$  \\
 & \textcolor{oursemph}{+ CCTC, core then confirm} & 64.0 & 5.0 & $92.1\%$ & 491.1 & 16.15$\times$ & 70.0 & 4.7 & $92.6\%$ & 386.4 & 13.70$\times$  \\
\rowcolor{finalres}
 & \textcolor{oursemph}{$\mathrm{C}^4$: CCTC on every block} & 64.0 & 4.8 & $92.5\%$ & 490.4 & 16.13$\times$ & 70.0 & 4.5 & $93.0\%$ & 389.7 & 13.82$\times$  \\
\midrule
\multicolumn{12}{l}{\textit{Long-reasoning tasks (multi-step CoT)}} \\
\midrule
GSM8K     & Baseline & 66.0 & 256.0 & $0.0\%$ & 19.0 & 1.00$\times$ & 85.0 & 256.0 & $0.0\%$ & 21.2 & 1.00$\times$ \\
(n=100, sequence length=256)& + confidence gate only & 61.0 & 59.2 & $76.9\%$ & 20.7 & 1.09$\times$ & 83.0 & 53.8 & $79.0\%$ & 21.3 & 1.00$\times$  \\
(block=32, 8 blocks)& + stability gate only & 33.0 & 24.5 & $90.4\%$ & 50.0 & 2.63$\times$ & 83.0 & 53.2 & $79.2\%$ & 23.3 & 1.10$\times$  \\
 & + CVEE (joint gate) & 66.0 & 250.0 & $2.3\%$ & 19.2 & 1.01$\times$ & 85.0 & 256.0 & $0.0\%$ & 21.1 & 1.00$\times$ \\
 & \textcolor{oursemph}{+ CCTC, core only} & 66.0 & 85.2 & $66.7\%$ & 51.8 & 2.73$\times$ & 85.0 & 85.8 & $66.5\%$ & 63.7 & 3.00$\times$  \\
 & \textcolor{oursemph}{+ CCTC, core then confirm} & 66.0 & 83.9 & $67.2\%$ & 52.0 & 2.74$\times$ & 85.0 & 84.3 & $67.1\%$ & 64.6 & 3.05$\times$  \\
\rowcolor{finalres}
 & \textcolor{oursemph}{$\mathrm{C}^4$: CCTC on every block} & 64.0 & 62.5 & $75.6\%$ & 67.5 & 3.55$\times$ & 85.0 & 54.5 & $78.7\%$ & 102.1 & 4.82$\times$  \\
\midrule
\multicolumn{12}{l}{\textit{Code generation (whole-program answer)}} \\
\midrule
HumanEval & Baseline & 40.2 & 384.0 & $0.0\%$ & 13.8 & 1.00$\times$ & 45.1 & 384.0 & $0.0\%$ & 14.5 & 1.00$\times$ \\
(n=164, sequence length=384)& + confidence gate only & 4.9 & 9.7 & $97.5\%$ & 67.6 & 4.90$\times$ & 5.5 & 9.5 & $97.5\%$ & 196.5 & 13.55$\times$  \\
(block=32, 12 blocks)& + stability gate only & 42.1 & 39.4 & $89.7\%$ & 24.8 & 1.80$\times$ & 43.3 & 29.2 & $92.4\%$ & 43.9 & 3.03$\times$  \\
 & + CVEE (joint gate) & 37.2 & 212.4 & $44.7\%$ & 24.8 & 1.80$\times$ & 42.7 & 109.2 & $71.6\%$ & 39.3 & 2.71$\times$ \\
 & \textcolor{oursemph}{+ CCTC, core only} & 40.9 & 50.0 & $87.0\%$ & 90.8 & 6.58$\times$ & 43.3 & 36.6 & $90.5\%$ & 116.5 & 8.03$\times$  \\
 & \textcolor{oursemph}{+ CCTC, core then confirm} & 42.1 & 48.8 & $87.3\%$ & 92.6 & 6.71$\times$ & 43.3 & 35.8 & $90.7\%$ & 119.0 & 8.21$\times$  \\
\rowcolor{finalres}
 & \textcolor{oursemph}{$\mathrm{C}^4$: CCTC on every block} & 42.1 & 39.4 & $89.7\%$ & 120.0 & 8.70$\times$ & 43.3 & 29.2 & $92.4\%$ & 146.0 & 10.07$\times$  \\
\bottomrule
\end{tabular}%
}
\caption{\textbf{Complementary savings from early exit and local commitment.}
Early stabilization makes CVEE the main source of savings on MMLU. On GSM8K and
HumanEval, CCTC further reduces computation within the blocks that generation still needs.
Highlighted rows show $\mathrm{C}^4$.}
\label{tab:blockaccel-ablation}\vspace{-2mm}
\end{table}

\vspace{-1mm}
\noindent\textbf{Confidence and agreement play complementary roles.}
The individual CVEE gates can stop quickly without consistently retaining task performance
(Table~\ref{tab:blockaccel-ablation}). Confidence alone performs poorly on HumanEval, while
stability alone sharply reduces accuracy on LLaDA GSM8K. By combining the signals, the joint
gate retains
full-decoding GSM8K accuracy on both backbones, but consumes nearly the full budget.
The joint gate therefore leaves most reasoning acceleration to local commitment rather than
obtaining it through earlier termination.

\vspace{-1mm}
\noindent\textbf{Structured commitment supplies the remaining acceleration.}
With the joint stopping gate in place, adding the CCTC core substantially reduces steps on
GSM8K and HumanEval. Building on this reduction, one-step confirmation provides a smaller, consistent further reduction across the reported tasks and
backbones; on HumanEval, it also improves the LLaDA point estimate while leaving Dream
accuracy unchanged. Furthermore, extending CCTC to the final block adds savings, with a score
trade-off on LLaDA GSM8K. The per-block profiles show the resulting distribution of work:
short-answer savings largely remove later blocks, whereas reasoning and code savings reduce
updates within blocks that remain active (Appendix~\ref{app:taublk}).

\vspace{-1mm}
\noindent\textbf{Scaling and caching.}
These component-level gains also depend on the decoding configuration. As the generation budget grows, $\mathrm{C}^4$'s executed steps grow more slowly, increasing
the fraction of computation saved. Block length instead changes the scope of parallel updates
and the prediction trajectory. Caching can further reduce runtime in some configurations,
with task-performance trade-offs. Appendix~\ref{app:scaling-cache} reports the full results.

%% file: sections/conclusion.tex
\vspace{-2mm}
\section{Conclusion}
\label{sec:conclusion}
\vspace{-2mm}

We presented $\mathrm{C}^4$, a training-free framework that reduces diffusion decoding cost
through local token commitment and adaptive global stopping. Its central design principle is
to match each decision to the evidence it needs: CCTC structures parallel commitments while
generation continues, and CVEE uses confidence and revision history to determine when to stop.
The experiments show why both controls matter. Early stabilization leaves entire blocks to
skip, whereas extended reasoning and code generation benefit more from reducing the work
inside blocks that still need to run. Reusing one numerical configuration across the evaluated
tasks and both backbones further demonstrates the practical value of adapting decisions to
prediction behavior without task-specific recalibration. Future work will extend this approach
to multimodal models and longer generations, and combine adaptive stopping with refinement that
preserves the coherence and completeness of the full response.

%% file: sections/appendix/overview.tex
\section*{Appendix Overview}

The appendices document the evaluation protocol and provide additional analyses of decoding behavior.
\begin{itemize}[leftmargin=*,itemsep=1pt,topsep=3pt]
\item \textbf{\ref{app:hyperparams}} Experimental settings and implementation details.
\item \textbf{\ref{app:retiming}} Decoding-loop timing protocol.
\item \textbf{\ref{app:span-control}} Commitment location and stopping behavior.
\item \textbf{\ref{app:cvee-trace}} Stopping decisions on real decoding trajectories.
\item \textbf{\ref{app:third-party-composition}} CVEE with WINO.
\item \textbf{\ref{app:taublk}} Per-block computation and threshold sensitivity.
\item \textbf{\ref{app:scaling-cache}} Generation length, block length, and caching.
\item \textbf{\ref{app:fill-quality}} Text quality after early exit.
\item \textbf{\ref{app:limitations}} Limitations and future directions.
\end{itemize}

%% file: sections/appendix/hyperparams.tex
\section{Experimental Setup and Decoding Settings}
\label{app:hyperparams}
\label{app:hyperparams-harness}

\paragraph{Prompting and evaluation protocol.}
All tasks use zero-shot prompts. Reasoning prompts add
``Let's think step by step,'' without demonstrations or additional answer-format instructions.
Dream-7B-Instruct uses its official one-position logit shift for next-token indexing; otherwise,
its evaluation protocol matches LLaDA-8B-Instruct. Prophet uses its published stopping
rule within the same prompting and evaluation pipeline.

\paragraph{Decoding parameters.}
Table~\ref{tab:hyperparams} lists the evaluation sizes, decoding budgets, and task-specific
settings. Across tasks and backbones, CVEE uses confidence threshold
$\tau_{\mathrm{CVEE}}{=}0.7$, revision multiplier $\gamma{=}2.0$, and minimum patience
$p_{\min}{=}3$. CCTC uses confidence threshold $\tau_{\mathrm{CCTC}}{=}0.8$ and gap
budget $H{=}1$. The extraction mode follows the task's output structure, as described below.
The trajectories in Figure~\ref{fig:cvee-trace} are excluded from CVEE calibration.

\begin{table}[H]
\centering
\small
\resizebox{\textwidth}{!}{%
\begin{tabular}{lccccccccc}
\toprule
Task & $n$ & Len. & Steps & Block & search\_mode &
Prophet ($\tau_{\text{high}}/\tau_{\text{mid}}/\tau_{\text{low}}$) &
SchED (mode/patience) &
KLASS ($\tau/\epsilon_{\text{KL}}$) &
WINO ($\tau/\tau_{\text{back}}$) \\
\midrule
MMLU       & $200$ & $64$  & $64$  & $16$ & first5 & $7.5/5.0/2.5$ & cosine / $0$ & $0.7/0.015$ & $0.5/0.9$ \\
ARC-C      & $200$ & $64$  & $64$  & $16$ & first5 & $7.5/5.0/2.5$ & cosine / $0$ & $0.7/0.015$ & $0.5/0.9$ \\
HellaSwag  & $200$ & $64$  & $64$  & $16$ & first5 & $7.5/5.0/2.5$ & cosine / $0$ & $0.7/0.015$ & $0.5/0.9$ \\
WinoGrande & $200$ & $64$  & $64$  & $16$ & first5 & $7.5/5.0/2.5$ & cosine / $0$ & $0.7/0.015$ & $0.5/0.9$ \\
PIQA       & $200$ & $64$  & $64$  & $16$ & first5 & $7.5/5.0/2.5$ & cosine / $0$ & $0.7/0.015$ & $0.5/0.9$ \\
\midrule
GSM8K      & $100$ & $256$ & $256$ & $32$ & last & $8.0/5.0/3.5$ & linear / $4$ & $0.6/0.015$ & $0.6/0.9$ \\
MATH       & $100$ & $256$ & $256$ & $32$ & last & $8.0/5.0/3.5$ & linear / $4$ & $0.6/0.015$ & $0.7/0.9$ \\
SVAMP      & $100$ & $256$ & $256$ & $32$ & last & $8.0/5.0/3.5$ & linear / $4$ & $0.6/0.015$ & $0.6/0.9$ \\
ASDiv      & $100$ & $256$ & $256$ & $32$ & last & $8.0/5.0/3.5$ & linear / $4$ & $0.6/0.015$ & $0.6/0.9$ \\
GSM-Hard   & $100$ & $256$ & $256$ & $32$ & last & $8.0/5.0/3.5$ & linear / $4$ & $0.6/0.015$ & $0.6/0.9$ \\
\midrule
HumanEval  & $164$ & $384$ & $384$ & $32$ & code & $7.5/5.0/4.5$ & cosine / $4$ & $0.9/0.01$ & $0.7/0.9$ \\
MBPP       & $257$ & $384$ & $384$ & $32$ & code & $7.5/5.0/4.5$ & cosine / $4$ & $0.7/0.01$ & $0.8/0.9$ \\
\bottomrule
\end{tabular}%
}
\caption{\textbf{Task settings used in Table~\ref{tab:main}.}
$n$ is the number of evaluated examples, Len. is the generation length, Steps is the full decoding
budget, and Block is the block length. \texttt{search\_mode} is determined by the expected output
format. KLASS code settings are for LLaDA; Dream values appear below.}
\label{tab:hyperparams}
\end{table}

\paragraph{Baseline settings.}
Prophet uses its published staged confidence thresholds, with a separate schedule for code
generation. SchED combines its calibrated decay schedule and patience with Prophet's high
and low thresholds. Its code experiments use the long-form stability guard, following the
available configuration for extended generation; SchED provides no dedicated code setting.

KLASS uses calibrated settings for each task family. For code generation, the released
$(\tau,\epsilon_{\mathrm{KL}})$ values are $(0.9,0.01)$ and $(0.7,0.01)$ for LLaDA on
HumanEval and MBPP, respectively; Dream uses $(0.8,0.001)$ and $(0.9,0.001)$.
WINO uses published task thresholds where available and the nearest task-family setting
otherwise, with $\tau_{\mathrm{back}}{=}0.9$. Fast-dLLM combines DualCache with threshold $0.9$.

\paragraph{Locating the monitored span.}
\label{app:hyperparams-searchmode}
At each step, a deterministic parser locates the span used by CVEE. The
\texttt{search\_mode} setting connects this extraction to the task's output structure.
For reasoning tasks (GSM8K, MATH, SVAMP, ASDiv, and GSM-Hard), \texttt{last} searches
the final $30\%$ of the generation (\texttt{tail\_frac}$=0.3$). For multiple-choice tasks
(MMLU, ARC-C, HellaSwag, WinoGrande, and PIQA), \texttt{first5} extracts the first answer
letter and locates its matching token sequence within the first five output positions.
For HumanEval and MBPP, \texttt{code} tracks the function body using Python indentation
to identify its end. Character spans are mapped directly to token indices to preserve
multi-line code boundaries.

\paragraph{Confidence and observation updates.}
CVEE observes spans within active or previously decoded blocks; positions beyond the active
block receive confidence $-\infty$. Code stability counters update when span confidence is
finite. Missing extractions pause the counters, so the agreement run counts successive
available observations, which can be separated by decoding steps with no extraction.

Each forward pass supplies the top-1 probabilities
$p_i^{\max}(t)=\max_v p_\theta(v\mid x^{(t)})_i$. CVEE averages these over the eligible
span $A_t$ to obtain $c_t=|A_t|^{-1}\sum_{i\in A_t}p_i^{\max}(t)$.
The extracted text combines current predictions at masked positions with retained tokens at
committed positions. At a committed position, the current top-1 prediction can differ from
the retained token, so its confidence does not independently verify that token.
When CVEE exits, the current predictions fill all remaining masks in the sequence.
This final fill reuses the existing forward pass.

%% file: sections/appendix/retiming.tex
\section{Throughput Measurement}
\label{app:retiming}

\paragraph{Timing scope.}
The RTX~4090 measurements cover the complete decoding loop, including model forward passes,
token updates, and method-specific operations such as extraction and stopping checks.
Before and after each timed decode, \texttt{torch.cuda.synchronize()} ensures that GPU work
is complete at the timing boundaries. Elapsed host time is measured with \texttt{time.time()}.
This procedure applies to Baseline, Prophet~\cite{li2026diffusion}, SlowFast Sampling
\citep{wei2025slowfast}, SchED~\citep{mohamed2025sched}, KLASS~\citep{kim2025klass},
WINO~\citep{hong2026wino}, and $\mathrm{C}^4$ in the main experiments and ablations.

\paragraph{Evaluation runs.}
Each method decodes the same fixed set of $20$ evaluation examples after warming up on two
additional examples. Three complete runs over the timing set yield the mean throughput
reported in the paper. Runs use batch size $1$, \texttt{bf16}, and greedy decoding on an idle
RTX~4090. Table~\ref{tab:tps-protocol} summarizes the environment and variation across runs.

\begin{table}[H]
\centering
\footnotesize
\renewcommand{\arraystretch}{0.95}
\begin{tabular}{lp{0.68\textwidth}}
\toprule
Setting & Value \\
\midrule
GPU & NVIDIA GeForce RTX 4090 \\
Batch size & $1$ \\
Precision & \texttt{bf16} \\
PyTorch / CUDA & $2.5.1$+cu121 / $12.1$ \\
Synchronization & \texttt{torch.cuda.synchronize()} before and after each timed decode \\
Timing & Host wall-clock time with \texttt{time.time()} \\
Examples / warmup / repeats & $20$ / $2$ / $3$ \\
LLaDA timing CV & $0.00$--$2.98\%$ (mean $0.14\%$) \\
Dream timing CV & $0.00$--$0.48\%$ (mean $0.08\%$) \\
\bottomrule
\end{tabular}
\caption{\textbf{RTX~4090 timing protocol.}
CV denotes the coefficient of variation across the three repeated timing runs for each measured
configuration.}
\label{tab:tps-protocol}
\end{table}

\paragraph{Reading the efficiency metrics.}
Step counts the executed decoding updates, TPS measures decoding throughput, and Speedup
is the ratio of method TPS to the corresponding full-decoding baseline TPS. Together, these
metrics describe how much decoding work is removed and how that reduction translates into
elapsed time, including the cost of each method's token selection and stopping operations.

%% file: sections/appendix/span_control.tex
\section{Commitment Location and Stopping Behavior}
\label{app:span-control}

Local updates shape both the context for future predictions and the observations available to
CVEE. We vary where CCTC may commit while retaining the reference top-$k$ schedule, separating
changes in the monitored span from changes in its surrounding context.

\paragraph{Effect of Protecting the Candidate Span.}
We first prevent CCTC from making additional commits inside the candidate span observed by CVEE.
Table~\ref{tab:span-control} reports four diagnostics. \emph{Frozen} is the fraction of the accepted
span already committed when CVEE fires, and \emph{By CCTC} is the fraction committed specifically
by CCTC rather than by the reference decoder's scheduled top-$k$ rule.
\emph{Recon.\ tok.} and \emph{Recon.\ exact} re-mask the accepted span on a copy of the current
buffer and measure whether one additional forward pass reconstructs the same tokens or the entire
span, respectively. \emph{Same ans.} is the fraction of examples whose graded answer remains
unchanged when CCTC is barred from committing inside the candidate span.

\begin{table}[h]
\centering
\small
\resizebox{\textwidth}{!}{%
\begin{tabular}{lrrrrrrrrrr}
\toprule
& \multicolumn{5}{c}{LLaDA-8B-Instruct}
& \multicolumn{5}{c}{Dream-7B-Instruct} \\
\cmidrule(lr){2-6}\cmidrule(lr){7-11}
Task
& Frozen & By CCTC & Recon.\ tok.\ & Recon.\ exact & Same ans.\
& Frozen & By CCTC & Recon.\ tok.\ & Recon.\ exact & Same ans.\ \\
\midrule
MMLU
& $68.1\%$ & $37.7\%$ & $97.9\%$ & $97.9\%$ & $100.0\%$
& $74.6\%$ & $50.3\%$ & $97.9\%$ & $97.9\%$ & $100.0\%$ \\
GSM8K
& $11.5\%$ & $7.7\%$ & $100.0\%$ & $100.0\%$ & $100.0\%$
& $0.0\%$ & $0.0\%$ & $100.0\%$ & $100.0\%$ & $100.0\%$ \\
HumanEval
& $99.9\%$ & $86.3\%$ & $33.8\%$ & $6.8\%$ & $59.1\%$
& $99.7\%$ & $75.1\%$ & $32.2\%$ & $12.5\%$ & $72.0\%$ \\
\bottomrule
\end{tabular}%
}
\caption{\textbf{Effect of protecting the candidate span from CCTC commits.}
For MMLU and GSM8K, removing CCTC's additional commits inside the candidate span leaves every
graded answer unchanged. HumanEval differs because the candidate is the generated program itself.}
\label{tab:span-control}
\end{table}

On MMLU and GSM8K, span protection changes stopping time slightly while preserving every
graded output on both backbones. High reconstruction agreement provides a second view of the
same behavior: the selected values remain supported by the current context after remasking.
These observations distinguish the timing of acceptance from the identity of the accepted output
in this control; they do not establish independence between commitment and stopping.

\paragraph{Effect of Protecting the Entire Candidate Block.}
Protecting only the candidate span still allows CCTC to commit neighboring positions in the same
block, which could indirectly affect the candidate trajectory. We therefore use a stricter control:
from the first step at which a candidate is extractable until CVEE fires, CCTC makes no additional
commits anywhere in the block containing that candidate. All other blocks retain the deployed rule.

Block protection delays exit more than span protection, indicating that updates to neighboring
positions help the monitored prediction settle. Most graded outputs remain unchanged, while the
few correctness changes occur in both directions (four gains and two losses; paired sign-test
$p=0.69$). The contrast between these controls highlights context development as part of the
interaction: local acceleration can advance stopping even when it does not directly commit the
monitored positions.

\paragraph{Effect on Whole-Program Answers.}
For HumanEval, the monitored unit covers the generated function, so protecting it suppresses
local acceleration over most of the program. The resulting increase in decoding cost is therefore
an effect of changing the generation process itself. LLaDA accuracy also decreases under this
control, while Dream accuracy stays unchanged.

Reconstruction behaves differently here as well. Rebuilding a long program in one parallel pass
requires recovering many interdependent tokens, and agreement remains low on both the deployed
and protected trajectories. The probe is thus sensitive to reconstruction difficulty as well as
to the state of the original generation. Whole-program protection and reconstruction cannot
separate local commitment from global stopping as cleanly as the short-span controls.

%% file: sections/appendix/cvee_trace.tex
\section{Prediction Evolution and Stopping Decisions}
\label{app:cvee-trace}
\renewcommand{\topfraction}{0.95}
\renewcommand{\textfraction}{0.05}
\renewcommand{\floatpagefraction}{0.85}

\begin{figure}[h]
\centering
\includegraphics[width=\linewidth]{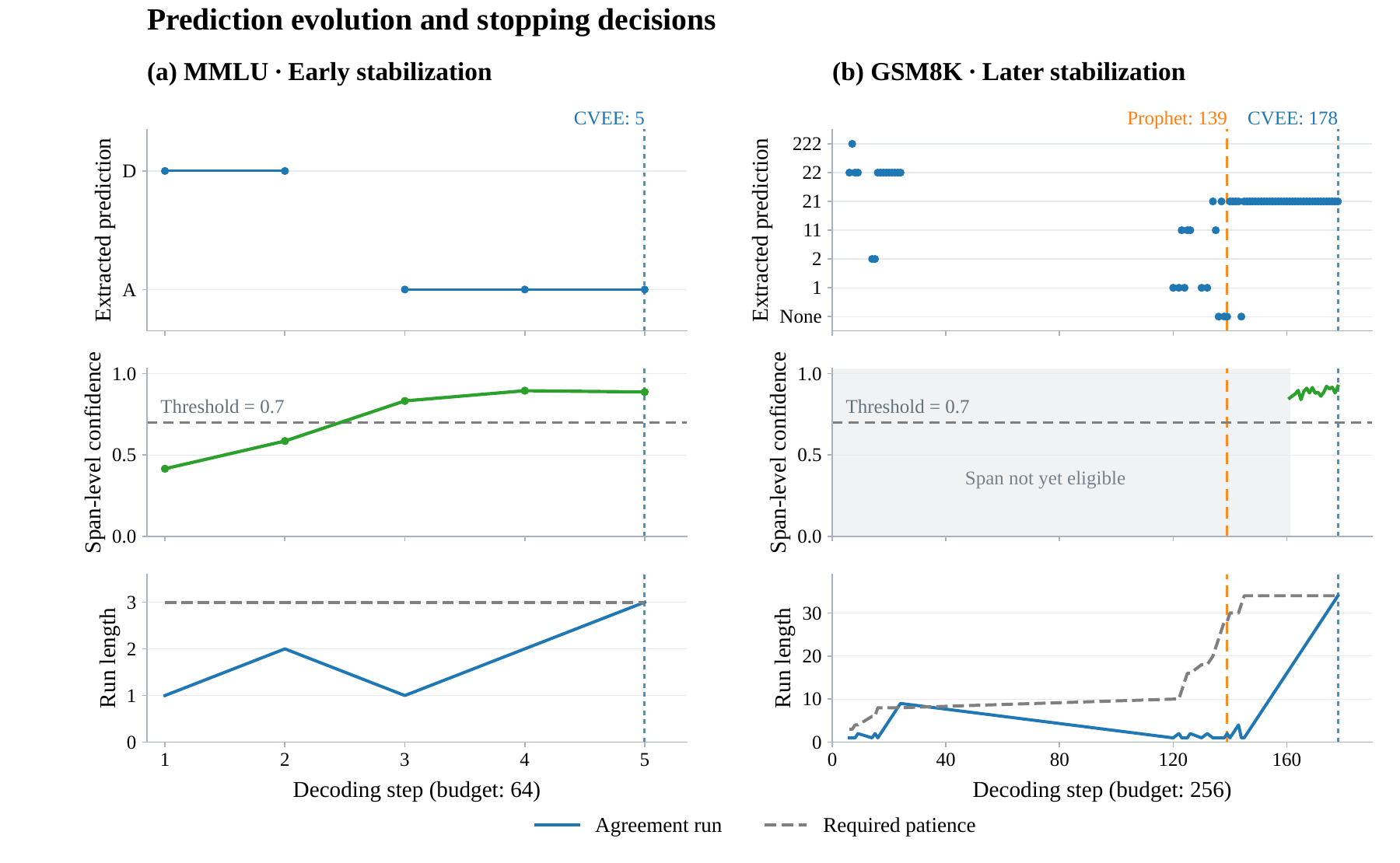}
\caption{\textbf{Early and late stabilization along individual decoding trajectories.}
Each column follows extracted predictions, span-level confidence, and temporal agreement.
MMLU settles early; GSM8K undergoes later revisions that increase the required patience.
Vertical markers locate CVEE's exit and, on the same GSM8K trajectory, Prophet's earlier exit.
Prediction values are categorical; the shaded region marks an ineligible span.}
\label{fig:cvee-trace-single}
\end{figure}

Figure~\ref{fig:cvee-trace-single} follows two trajectories through the signals used by CVEE.
Reading each column from top to bottom connects changes in the extracted prediction to
confidence and accumulated agreement. The examples complement the task-level patterns in
Figure~\ref{fig:cvee-trace} by showing how stopping decisions emerge over successive steps.

\paragraph{Early stabilization.}
On MMLU, an early revision from D to A is followed by rising confidence and repeated
agreement. Confidence crosses its threshold before the required run is complete; CVEE exits
at step 5 of 64 once both conditions hold. Most of the generation budget can therefore be
skipped after a short period of stable predictions.

\paragraph{Later stabilization.}
On GSM8K, the extracted value changes repeatedly before settling on 21. These revisions
increase the required patience, while confidence becomes available when the span is eligible.
CVEE exits at step 178 of 256 as the agreement run reaches that requirement. Prophet's exit
at step 139 is marked on the same trajectory, placing both stopping decisions in the context
of the evolving predictions. The longer revision history gives temporal agreement a larger
role in determining when this trajectory ends.

\paragraph{From stopping decisions to generated text.}
Figures~\ref{tab:case-study} and~\ref{tab:case-study-svamp} show the corresponding output-level
behavior. All displayed methods produce the correct numerical values, while their final
sentences differ in fluency and completion. Badges mark answer presence in the literal text.
The released extractor can match trailing \texttt{..} and return an empty string, so these
badges are assigned by inspecting the displayed outputs. Appendix~\ref{app:fill-quality} examines one-shot-fill quality across tasks for
$\mathrm{C}^4$ and the baselines.

\begin{figure}[h]
\centering
\begin{minipage}{\linewidth}
\begin{caseprompt}
\textbf{Prompt:} Liam and Mitchell own competing lemonade stands across the street from one another.
When Liam bragged that he had made \$63 one weekend, Mitchell laughed and told Liam he had sold 21
lemonades at \$4 apiece the same weekend. How many more dollars did Mitchell make selling lemonade
that weekend than Liam?
\end{caseprompt}

\begin{casepanel}{csbarplain}{csframe}{\csmodel{Baseline}{256/256 steps}}{\textbf{Answer present}}
To determine how many more dollars Mitchell made compared to Liam, we need to follow these steps:\csgap
\textbf{1. Calculate Mitchell's earnings:}\\
Mitchell sold 21 lemonades. Each lemonade was sold for \$4. Therefore:
\csdisp{21 x 4 = 84 dollars}
\textbf{2. Determine the difference:}\\
Liam made \$63. Mitchell made \$84.
\csdisp{84 - 63 = \csnum{21} dollars}
So, Mitchell made \$\csnum{21} more than Liam.
\end{casepanel}

\begin{casepanel}{csbarplain}{csframe}{\csmodel{Prophet}{139/256 steps}}{\textbf{Answer present}}
To determine how many more dollars Mitchell made compared to Liam, we need to follow these steps:\csgap
\textbf{1. Calculate Mitchell's earnings:}\\
Mitchell sold 21 lemonades. Each lemonade was sold for \$4. Therefore:
\csdisp{21 x 4 = 84 dollars}
\textbf{2. Determine the difference:}\\
Liam made \$63. Mitchell made \$84.
\csdisp{84 - 63 = \csnum{21} dollars}
So, Mitchell made \$\csnum{21} more \cserr{than than..}
\csnote{(correct numerical answer present; repeated words and malformed punctuation)}
\end{casepanel}

\begin{casepanel}{csbarhi}{csframehi}{\csmodel{\textcolor{csframehi}{$\mathrm{C}^4$}}{178/256 steps}}{\textbf{Answer present}}
To determine how many more dollars Mitchell made compared to Liam, we need to follow these steps:\csgap
\textbf{1. Calculate Mitchell's earnings:}\\
Mitchell sold 21 lemonades. Each lemonade was sold for \$4. Therefore:
\csdisp{21 x 4 = 84 dollars}
\textbf{2. Determine the difference:}\\
Liam made \$63. Mitchell made \$84.
\csdisp{84 - 63 = \csnum{21} dollars}
So, Mitchell made \$\csnum{21} more dollars than Liam.
\csnote{(committed here; budget left unspent)}
\end{casepanel}
\caption{\textbf{Literal decoded text at each method's stopping step.}
Real GSM8K trajectory with ground-truth answer $21$, corresponding to
Figure~\ref{fig:cvee-trace-single}. All outputs state the correct answer, highlighted in blue.
Badges indicate answer presence in the displayed text, not automated evaluator scores;
red text marks malformed wording or punctuation.}
\label{tab:case-study}
\end{minipage}
\end{figure}

\begin{figure}[h]
\centering
\begin{minipage}{\linewidth}
\begin{caseprompt}
\textbf{SVAMP (ground truth = 15)}
\end{caseprompt}

\begin{casepanel}{csbarplain}{csframe}{\csmodel{Baseline}{256/256 steps}}{\textbf{Answer present}}
Sure, let's break it down step by step:\csgap
\textbf{1. Initial number of cakes:}\\
The baker initially made 155 cakes.\\
\textbf{2. Cakes bought by the friend:}\\
The friend bought 140 cakes from the baker.\csgap
To find out how many cakes the baker still has, we need to subtract the number of cakes bought by
the friend from the initial number of cakes.
\csdisp{155 - 140 = \csnum{15}}
So, the baker still has \csnum{15} cakes.
\end{casepanel}

\begin{casepanel}{csbarplain}{csframe}{\csmodel{Prophet}{92/256 steps}}{\textbf{Answer present}}
Sure, let's break it down step by step:\csgap
\textbf{1. Initial number of cakes:}\\
The baker initially made 155 cakes.\\
\textbf{2. Cakes bought by the friend:}\\
The friend bought 140 cakes from the baker.\csgap
To find out how many cakes the baker still has, we need to subtract the number of cakes bought by
the friend from the initial number of cakes.
\csdisp{155 - 140 = \csnum{15}}
\cserr{SoSo, the would still still}\ \csnum{15}\ \cserr{cakes cakes.}
\csnote{(correct numerical answer present; repeated words and malformed final sentence)}
\end{casepanel}

\begin{casepanel}{csbarhi}{csframehi}{\csmodel{\textcolor{csframehi}{$\mathrm{C}^4$}}{256/256 steps}}{\textbf{Answer present}}
Sure, let's break it down step by step:\csgap
\textbf{1. Initial number of cakes:}\\
The baker initially made 155 cakes.\\
\textbf{2. Cakes bought by the friend:}\\
The friend bought 140 cakes from the baker.\csgap
To find out how many cakes the baker still has, we need to subtract the number of cakes bought by
the friend from the initial number of cakes.
\csdisp{155 - 140 = \csnum{15}}
So, the baker still has \csnum{15} cakes.
\csnote{(CVEE uses the full budget; the displayed answer and final sentence are correct)}
\end{casepanel}
\caption{\textbf{Representative SVAMP trajectory.}
All three outputs state the correct answer, $15$. Prophet stops earlier with a malformed
final sentence; CVEE uses the full budget on this example. This comparison illustrates text
quality rather than a difference in numerical-answer correctness.}
\label{tab:case-study-svamp}
\end{minipage}
\end{figure}

\begin{figure}[h]
\centering
\begin{minipage}{\linewidth}
\begin{caseprompt}
\textbf{PIQA (ground truth = A)}
\end{caseprompt}

\begin{casepanel}{csbarplain}{csframe}{\csmodel{Baseline}{64/64 steps}}{\csok}
A. Pile your dirty clothing items into a zip it compressor bag and place bottom of your suitcase.
\end{casepanel}

\begin{casepanel}{csbarplain}{csframe}{\csmodel{Prophet}{16/64 steps}}{\csok}
A. Pile packing dirty clothing items into zip it compressor bag and place in the bottom of your
suitcase.
\csnote{(the extracted letter is still correct, but the one-shot fill drops and reorders words
downstream of the answer token itself)}
\end{casepanel}

\begin{casepanel}{csbarhi}{csframehi}{\csmodel{\textcolor{csframehi}{$\mathrm{C}^4$}}{3/64 steps}}{\csok}
A. Pile your dirty clothing items into a zip it compressor bag and place in the bottom of your
suitcase.
\csnote{(committed after 3 steps; the rest of the buffer is still filled in by the ordinary
schedule, not frozen mid-word)}
\end{casepanel}
\caption{\textbf{Representative PIQA trajectory.}
The same literal-decoded-text comparison as Figure~\ref{tab:case-study}. Prophet's one-shot fill can degrade surrounding text even when the extracted answer is correct.}
\label{tab:case-study-piqa}
\end{minipage}
\end{figure}
\subsection{Consistency in Reasoning and Code}
\label{app:paired-cases}

Reasoning and code generation require dependencies to remain consistent across the output.
An intermediate quantity must carry through to the final calculation, while a computed condition
must lead to the requested program behavior. Figures~\ref{fig:reasoning-code-cases}
and~\ref{fig:code-return-case} illustrate these dependencies through selected GSM8K and
HumanEval examples from LLaDA-8B-Instruct.

\begin{figure}[h]
\centering
\begin{minipage}{\linewidth}
\begin{caseprompt}
\textbf{GSM8K: Collecting cards.} Elaine starts with 20 cards, collects 60 in the first month,
40 in the second, and twice the combined first- and second-month amount in the third.
The required total is \textbf{320}.
\end{caseprompt}

\begin{casepanel}{csbarplain}{csframe}{\csmodel{Prophet}{189/256 steps}}{\textbf{Answer present}}
So,{,} now has a total of \csnum{320} Pokemon cards.
\end{casepanel}

\begin{casepanel}{csbarplain}{csframe}{\csmodel{WINO}{44/256 steps}}{\textbf{Answer: 290}}
Second month:\\
\cserr{$60\div20=30$ cards}\\
Total: $20+60+30+180=290$\\
So, Elaine now has a total of **290** Pokemon cards.
\end{casepanel}

\begin{casepanel}{csbarplain}{csframe}{\csmodel{SlowFast}{35/256 steps}}{\textbf{Answer: 220}}
Third month: $2\times100=200$\\
Total: $20+60+40+\cserr{100}=220$\\
So, Elaine now has a total of 220 Pokemon cards.
\end{casepanel}

\begin{casepanel}{csbarhi}{csframehi}{\csmodel{\textcolor{csframehi}{$\mathrm{C}^4$}}{34/256 steps}}{\textbf{Answer present}}
Second month: $60-20=40$ cards\\
Third month: $2\times(60+40)=2\times100=200$ cards\\
Total: $20+60+40+200=\csnum{320}$\\
So, Elaine now has a total of \csnum{320} Pokemon cards.
\end{casepanel}
\caption{\textbf{Arithmetic consistency in a selected GSM8K example.}
Excerpts show intermediate calculations and final totals, with equations rendered for readability.
WINO changes the second-month calculation; SlowFast substitutes a different value in its final
sum. Prophet and $\mathrm{C}^4$ both state the correct total. Red highlights the inconsistent
calculations. All methods use LLaDA-8B-Instruct.}
\label{fig:reasoning-code-cases}
\end{minipage}
\end{figure}

\FloatBarrier
\paragraph{Carrying intermediate results forward.}
The card-counting example links each month's contribution to a final sum. Consistency can break
when a quantity is first computed or when it is reused: WINO changes the second-month amount,
whereas SlowFast obtains the third-month amount correctly but substitutes a different value
in the total. The $\mathrm{C}^4$ output preserves this chain of calculations. Prophet also reaches
the correct total despite malformed final wording, separating the arithmetic result from the fluency of its verbal expression.

\begin{figure}[h]
\centering
\begin{minipage}{\linewidth}
\begin{caseprompt}
\textbf{HumanEval: Returning $x$ or $y$.} Implement \texttt{x\_or\_y(n, x, y)}:
return $x$ when $n$ is prime and $y$ otherwise. For example,
\texttt{x\_or\_y(7, 34, 12)} returns 34 and \texttt{x\_or\_y(3, 33, 5212)} returns 33.
\end{caseprompt}

\begin{casepanel}{csbarplain}{csframe}{\csmodel{Prophet}{127/384 steps}}{\textbf{Syntax error}}
while i * i <= n:\\
\hspace*{2em}\cserr{if n \% i == 0 or n \%i + + == 0 ==}\\
\cserr{return return}
\end{casepanel}

\begin{casepanel}{csbarplain}{csframe}{\csmodel{WINO}{45/384 steps}}{\textbf{Fails tests}}
if n <= 1:\\
\hspace*{2em}\cserr{return False}\\
if n <= 3:\\
\hspace*{2em}\cserr{return True}
\end{casepanel}

\begin{casepanel}{csbarplain}{csframe}{\csmodel{SlowFast}{39/384 steps}}{\textbf{Fails tests}}
def x\_or\_y(n, x, y):\\
\hspace*{2em}\cserr{return x if n == 7 else y}
\end{casepanel}

\begin{casepanel}{csbarhi}{csframehi}{\csmodel{\textcolor{csframehi}{$\mathrm{C}^4$}}{25/384 steps}}{\textbf{Passes tests}}
if n <= 1:\\
\hspace*{2em}return y\\
if n <= 3:\\
\hspace*{2em}return x
\end{casepanel}
\caption{\textbf{Return behavior in a selected HumanEval example.}
Excerpts isolate the relevant branches of each generated program. Prophet's output contains
invalid syntax; WINO returns Boolean values; SlowFast handles only the visible prime example.
$\mathrm{C}^4$ returns the requested values. Test outcomes refer to the complete programs.}
\label{fig:code-return-case}
\end{minipage}
\end{figure}

\FloatBarrier
\paragraph{Connecting conditions to return values.}
In the code example, determining primality is an intermediate operation: the function must use
that condition to choose between $x$ and $y$. Returning a Boolean loses this connection, while
checking only $n=7$ replaces the general rule with a single example. The completed $\mathrm{C}^4$
program connects a general primality check to the requested return values and passes the recorded
tests. Together, these examples illustrate how correctness depends on carrying intermediate
computations through to their intended use.

%% file: sections/appendix/wino_cvee.tex
\section{Early Exit with WINO}
\label{app:third-party-composition}

We apply CVEE to WINO to examine stopping behavior on a different sampling trajectory
(Table~\ref{tab:composition}). WINO retains its local update rule, while CVEE evaluates
sequence termination using predictions and probabilities from the same forward passes.
This configuration pairs WINO commitments with CVEE stopping, so the local sampling trajectory is determined entirely by WINO.

\paragraph{Early stabilization leaves room for termination.}
On MMLU and HellaSwag, CVEE removes most of WINO's remaining steps with nearly unchanged
accuracy (Table~\ref{tab:composition}). Local sampling has accelerated token updates, but
stabilization still precedes completion of the scheduled generation. CVEE uses this interval to skip updates after the prediction has settled.

\paragraph{Effect on long-reasoning and code.}
The same composition is less effective on GSM8K, SVAMP, and HumanEval. With the fixed CVEE
configuration, accuracy drops by $23$--$36$ points. We additionally evaluate $45$ combinations of
$\tau_{\text{CVEE}} \in \{0.7,0.8,0.9\}$,
$p_{\min} \in \{3,5,8,12,16\}$, and
$\gamma \in \{2,4,8\}$ for each task. Recovering accuracy requires substantially more decoding.
The recovered operating points lie close to WINO's original step count. Thus, in these tasks,
the remaining trajectory contains little computation that this exit rule can remove while
retaining the original task performance.

\paragraph{Effect of WINO commitments on CVEE.}
The task contrast reflects where useful computation remains after local acceleration.
Reasoning trajectories continue changing during much of their shortened decoding horizon,
whereas short-answer trajectories can stabilize with several scheduled steps still available for CVEE to remove through global termination.

WINO's local commitments also affect the evidence observed by CVEE. Varying
$\tau_{\text{CVEE}}$ at fixed $(p_{\min},\gamma)$ does not change accuracy on the three
long-output tasks, indicating that confidence is not the limiting condition. This observation is consistent with the temporal condition determining the tested trade-off.
Because committed positions retain their tokens, repeated extractions can also reflect local
commitments; the threshold sweep alone does not identify the cause of premature exits.

This result illustrates the coupling between local commitment and global termination.
Appendix~\ref{app:span-control} examines the same interaction within $\mathrm{C}^4$.
CCTC instead structures additional commitments through a boundary-anchored core and one-step
confirmation alongside CVEE tracking of the answer.

%% file: sections/appendix/taublk.tex
\section{Computation Profiles and Commitment Thresholds}
\label{app:taublk}

\begin{figure}[h]
\centering
\includegraphics[width=\textwidth]{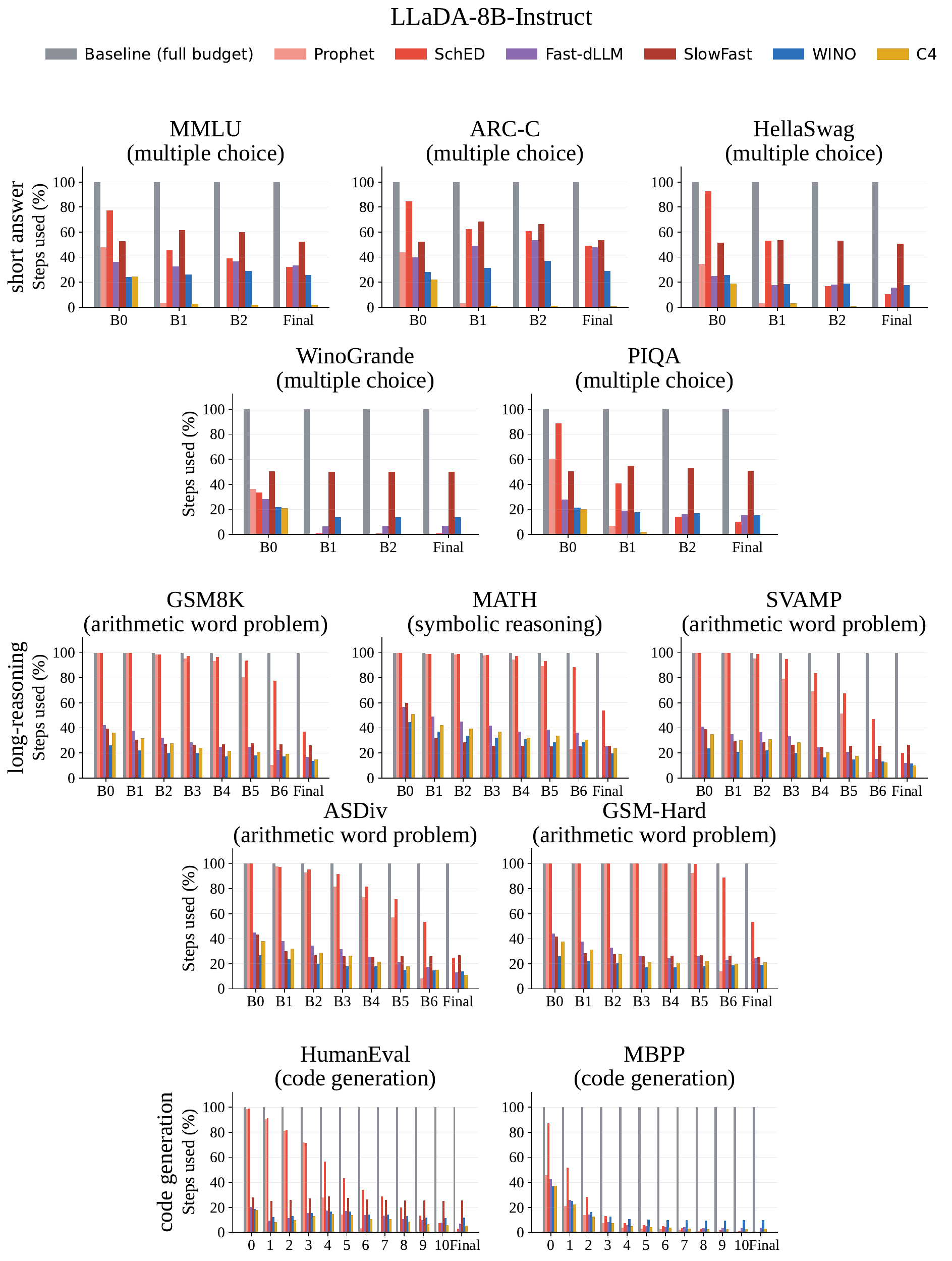}
\caption{\textbf{Per-block step usage across all $12$ tasks, LLaDA-8B-Instruct.}
Short-answer tasks terminate early, whereas long-reasoning and code tasks retain more
within-block computation.}
\label{fig:blockaccel-full-short}
\end{figure}

\begin{figure}[h]
\centering
\includegraphics[width=\textwidth]{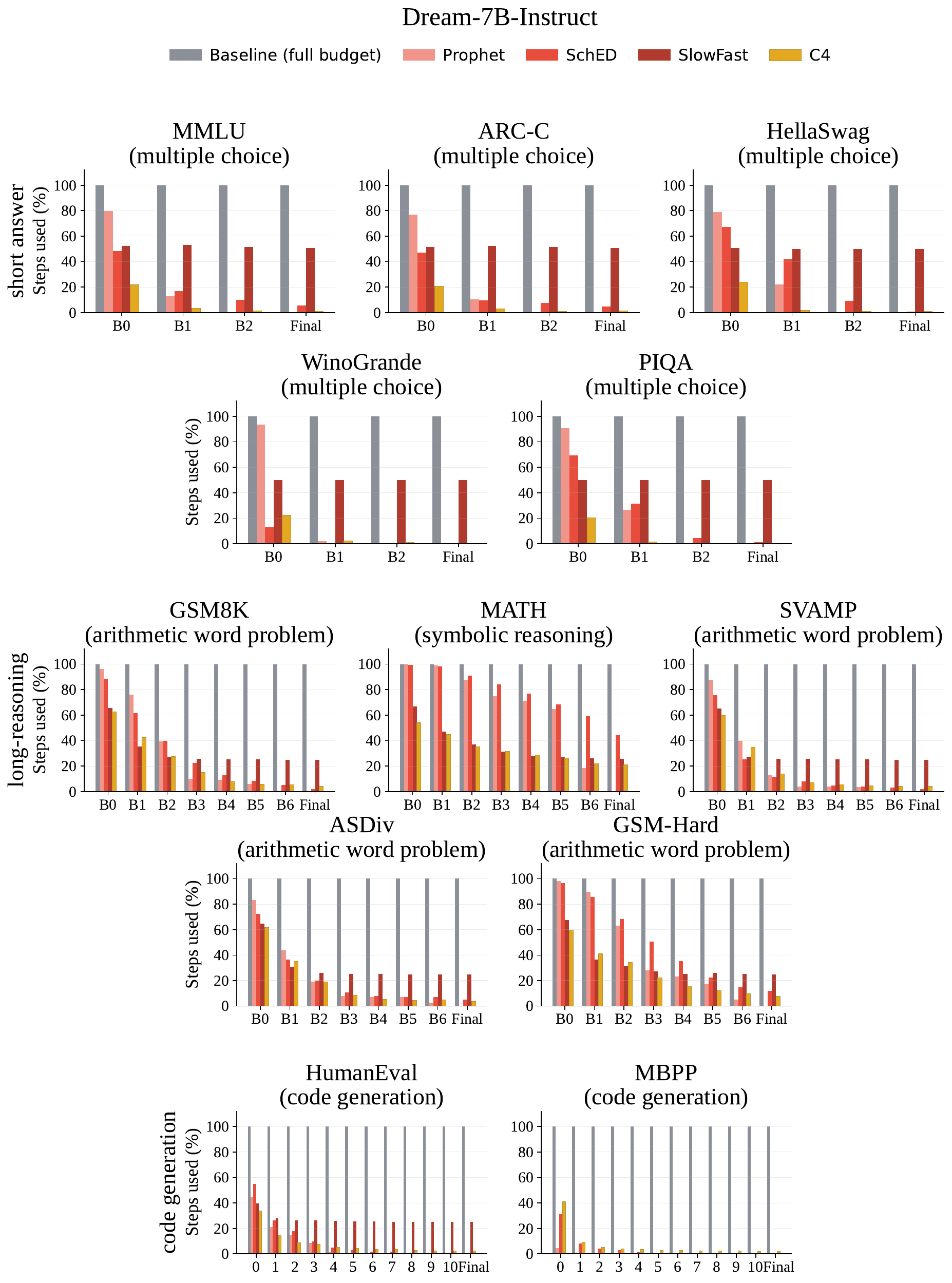}
\caption{\textbf{Per-block step usage across all $12$ tasks, Dream-7B-Instruct.}
Short-answer computation is concentrated in early blocks; reasoning and code extend across
more of the sequence, with savings from fewer updates within blocks.}
\label{fig:blockaccel-full-long}
\end{figure}

\paragraph{Where decoding work remains.}
The block profiles show how stabilization timing shapes computation across the sequence
(Figures~\ref{fig:blockaccel-full-short} and~\ref{fig:blockaccel-full-long}). Short-answer tasks
usually finish within the first one or two blocks, leaving later blocks unused. Reasoning and
code continue farther into the sequence, where CCTC reduces the updates within each block.
These patterns connect the component contributions in Section~\ref{sec:blockaccel-ablation}
to the location of the work saved during generation.

\begin{figure}[h]
\centering
\includegraphics[width=\textwidth]{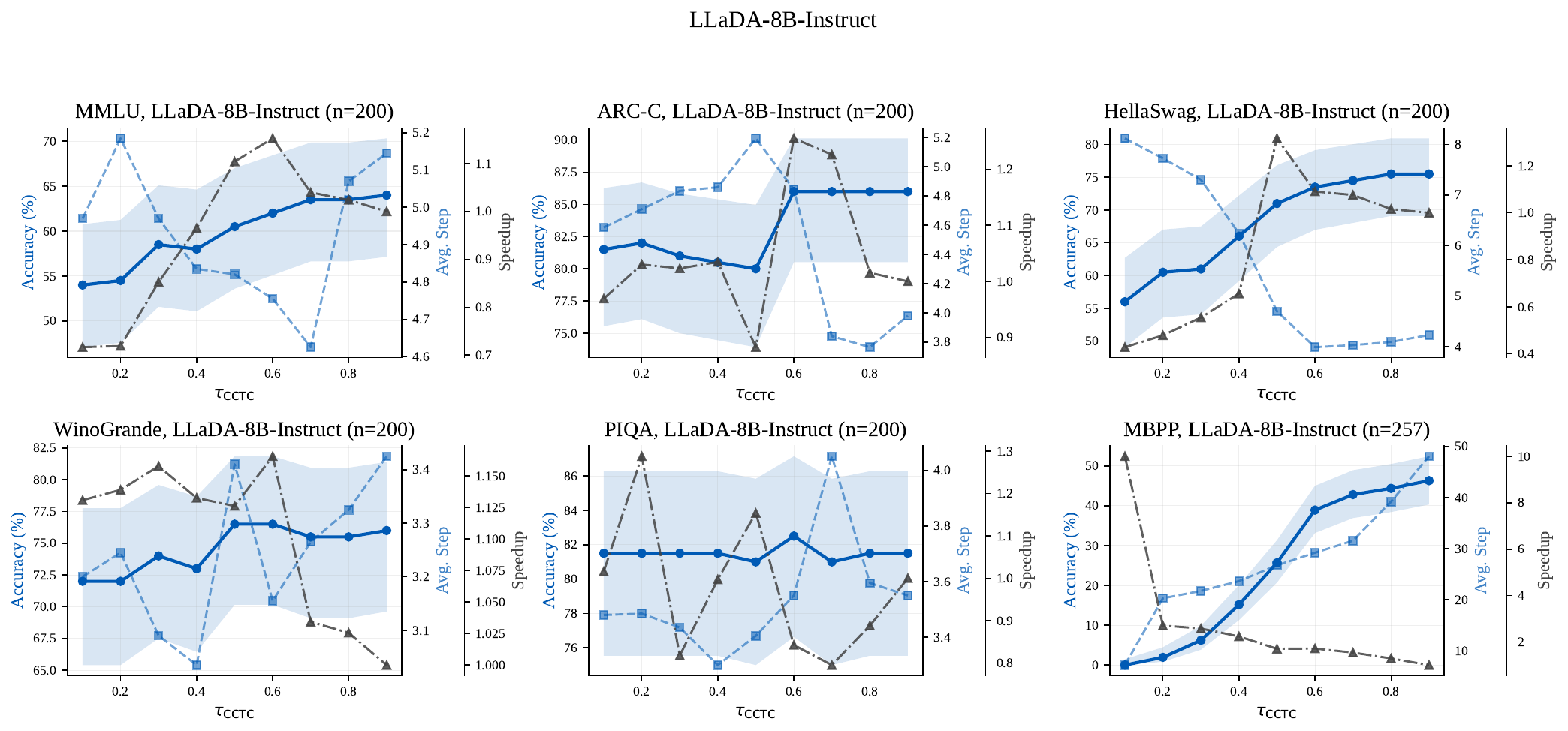}
\vspace{4pt}
\includegraphics[width=\textwidth]{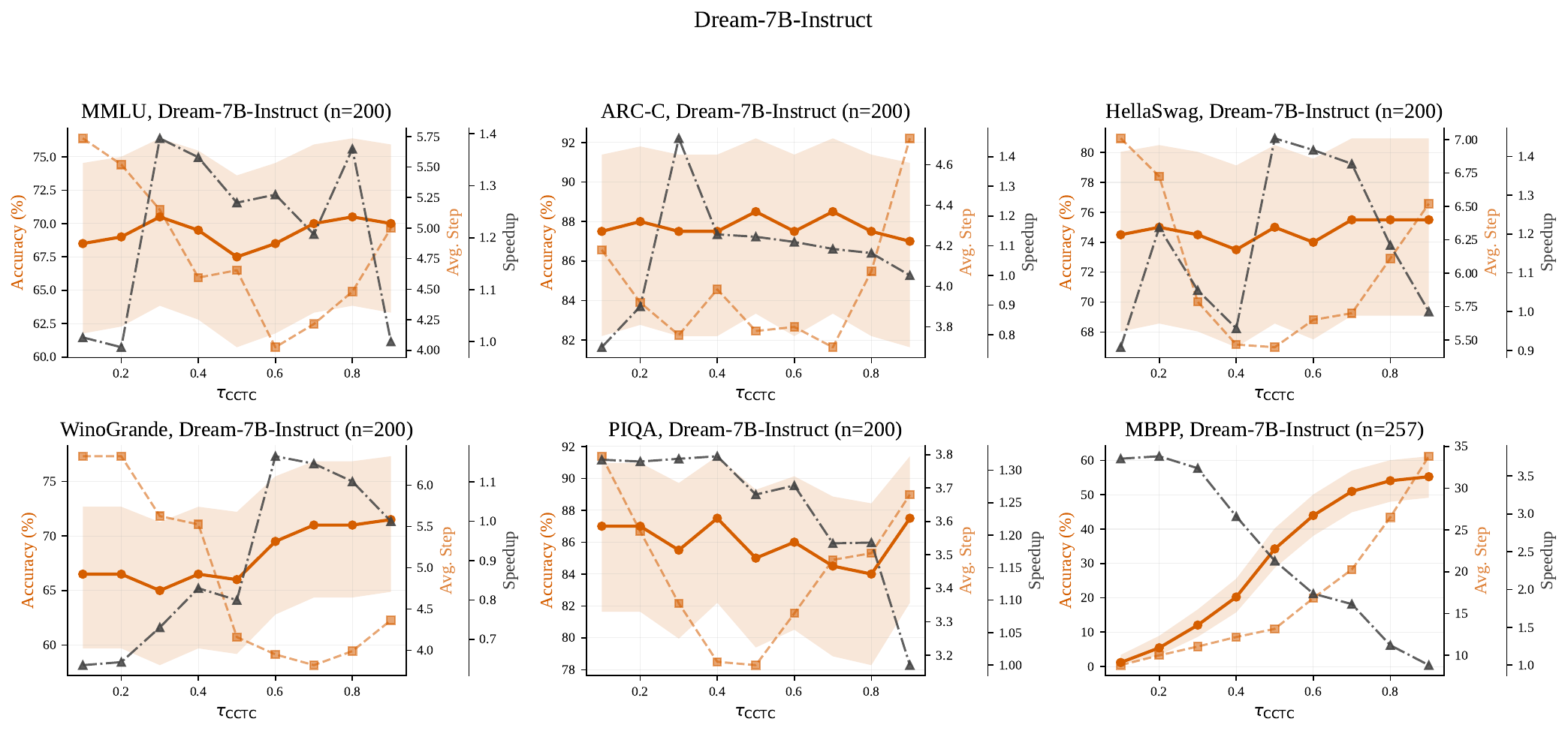}
\caption{\textbf{Effect of $\tau_{\mathrm{CCTC}}$ on short-answer tasks and MBPP.}
LLaDA-8B-Instruct is shown above and Dream-7B-Instruct below. Solid curves show accuracy
with Wilson-score $95\%$ intervals, dashed curves show average Step, and dash-dot curves show
Speedup relative to $\tau_{\mathrm{CCTC}}{=}0.9$. Accuracy remains stable across most of the
threshold range on the five short-answer tasks, while MBPP accuracy declines with more
aggressive commitment. Measurements use an NVIDIA A100.}
\label{fig:backfire-full-short}
\end{figure}

\begin{figure}[h]
\centering
\includegraphics[width=\textwidth]{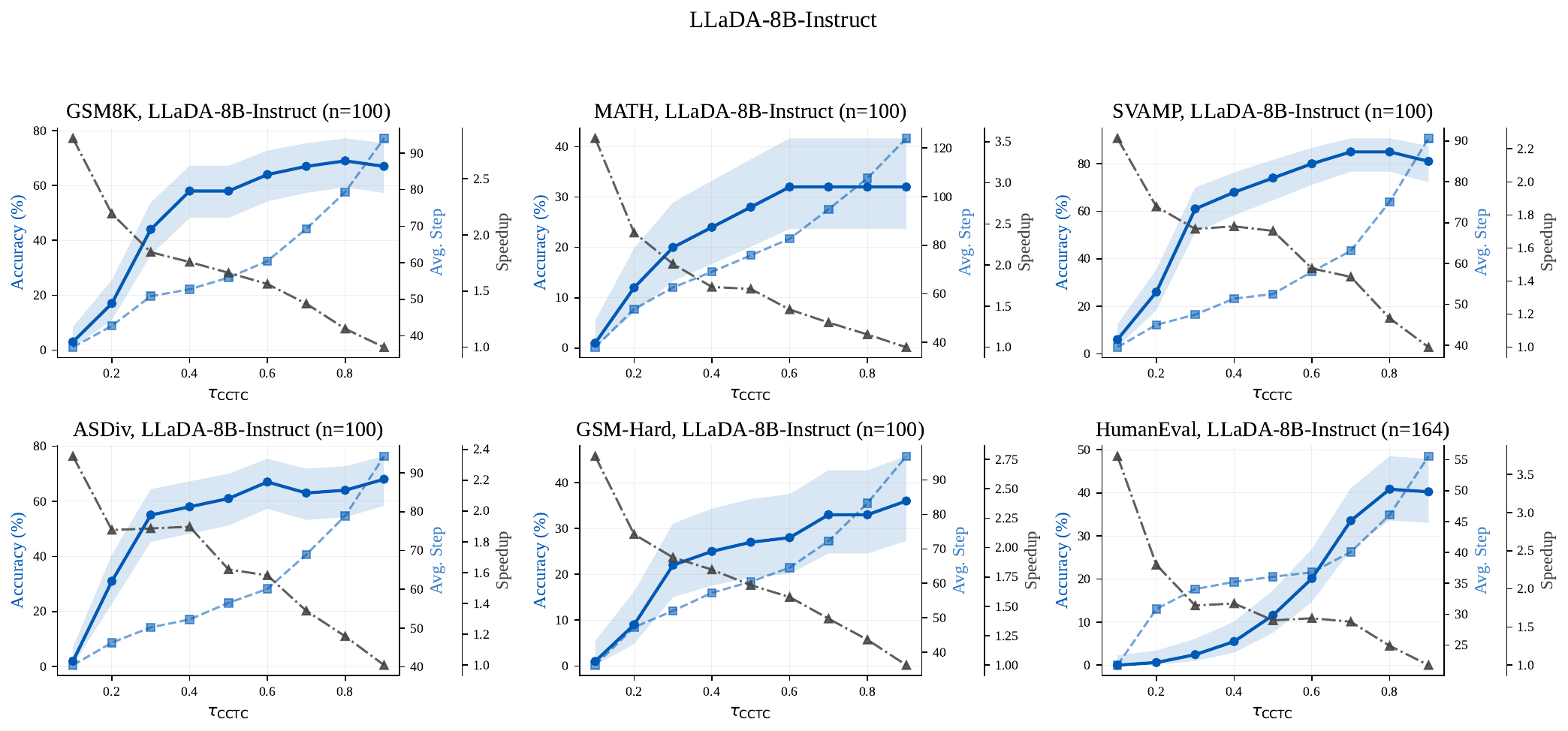}
\vspace{4pt}
\includegraphics[width=\textwidth]{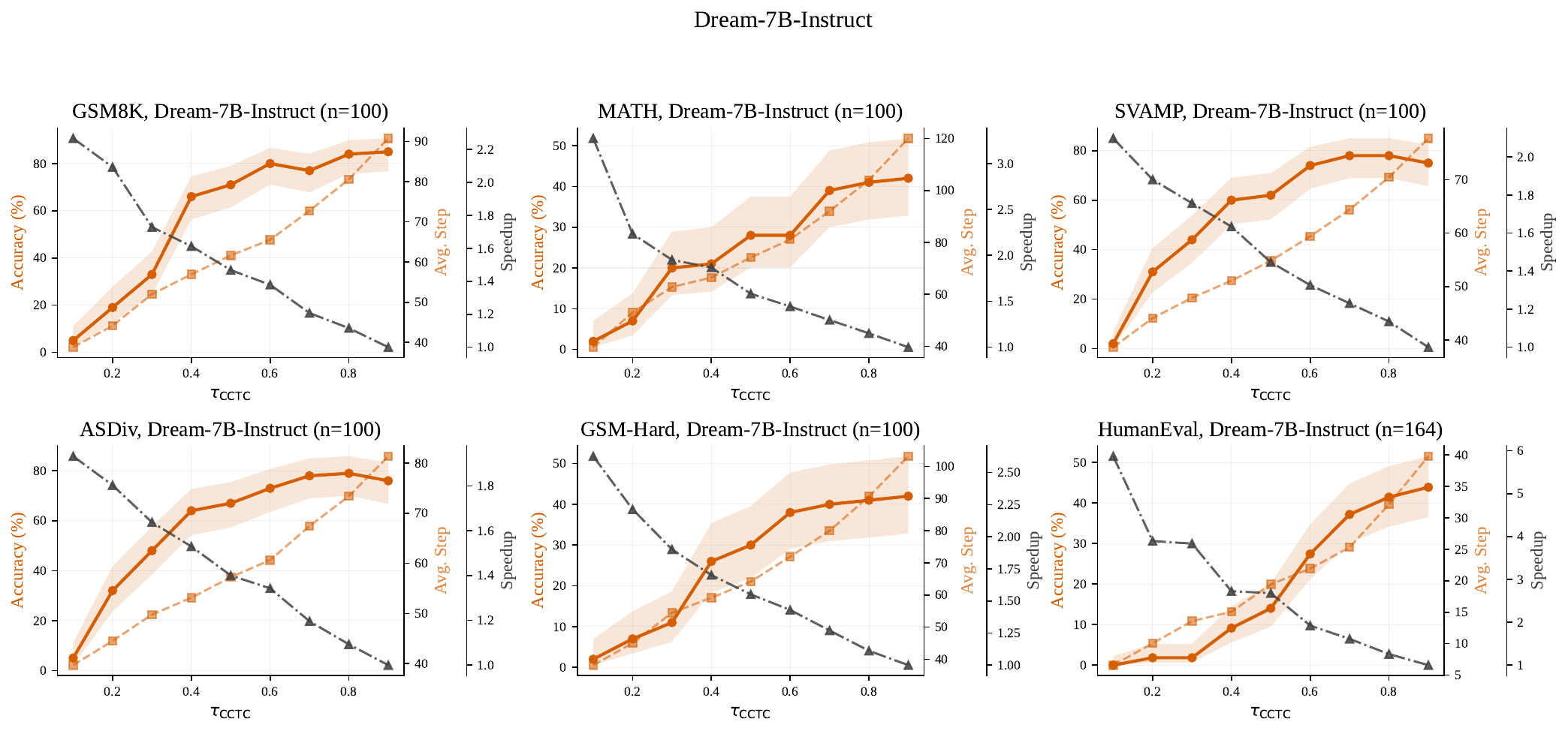}
\caption{\textbf{Effect of $\tau_{\mathrm{CCTC}}$ on long-reasoning tasks and HumanEval.}
Lower thresholds reduce decoding work but can sharply decrease accuracy, revealing a
quality cost to committing more predictions under incomplete context. Top: LLaDA-8B-Instruct. Bottom: Dream-7B-Instruct. Timings are measured on an NVIDIA A100 and normalized to the $0.9$ threshold setting.}
\label{fig:backfire-full-long}
\end{figure}

\paragraph{How the threshold shapes commitment.}
The confidence threshold $\tau_{\mathrm{CCTC}}$ controls which masked positions become eligible
for additional commitment. Lower values admit more predictions per update, reducing the
opportunity to revise them under later context. The sweeps in
Figures~\ref{fig:backfire-full-short} and~\ref{fig:backfire-full-long} trace the resulting changes
in accuracy and computation across the $12$ tasks; the main experiments use $0.8$ throughout.
Sweep speedups use NVIDIA A100 timings normalized to the $0.9$ setting.

The threshold has different consequences across task regimes. Short-answer accuracy changes
little over much of the tested range, while reasoning and code exhibit sharper declines as
commitment becomes more aggressive. Step counts and throughput change more gradually,
so a modest computational saving can carry a substantial accuracy cost. The useful operating
range therefore depends on how much a task benefits from revising intermediate predictions.

%% file: sections/appendix/scaling_cache.tex
\section{Generation Length, Block Length, and Caching}
\label{app:scaling-cache}

\begin{table}[!ht]
\centering
\footnotesize
\renewcommand{\arraystretch}{0.66}
\resizebox{\textwidth}{!}{%
\begin{tabular}{rrr c >{\columncolor{yellow!15}}r >{\columncolor{yellow!15}}r
                >{\columncolor{yellow!15}}r >{\columncolor{yellow!15}}r
                >{\columncolor{yellow!15}}r
                >{\columncolor{finalres}}r >{\columncolor{finalres}}r
                >{\columncolor{finalres}}r
                >{\columncolor{finalres}}r >{\columncolor{finalres}}r rrrrr}
\toprule
\multicolumn{4}{l}{\textbf{LLaDA-8B-Instruct}} & \multicolumn{5}{>{\columncolor{yellow!15}}c}{\textcolor{oursemph}{$\mathrm{C}^4$}} & \multicolumn{5}{>{\columncolor{finalres}}c}{$+$\,DualCache} & \multicolumn{5}{c}{Fast-dLLM (DualCache+parallel)} \\
\cmidrule(lr){5-9}\cmidrule(lr){10-14}\cmidrule(lr){15-19}
 $L$ & $B$ & $N$ & Baseline & Acc & Step & Saving & TPS & Speedup & Acc & Step & Saving & TPS & Speedup & Acc & Step & Saving & TPS & Speedup \\
\midrule
\multicolumn{19}{l}{\textit{HumanEval (code generation), block length swept at fixed generation length $384$}} \\
\midrule
 384 & 16 & 24 & $41.5$ & $39.6$ & $44.5$ & $88.4\%$ & 104.5 & 7.56$\times$ & $32.3$ & $45.4$ & $88.2\%$ & 127.1 & 9.22$\times$ & $39.0$ & $64.4$ & $83.2\%$ & 102.2 & 7.41$\times$ \\
 384 & 32 & 12 & $40.2$ & $42.1$ & $39.4$ & $89.7\%$ & 120.0 & 8.70$\times$ & $34.8$ & $44.7$ & $88.4\%$ & 154.3 & 11.18$\times$ & $37.8$ & $57.2$ & $85.1\%$ & 132.2 & 9.58$\times$ \\
 384 & 64 &  6 & $39.6$ & $42.1$ & $36.5$ & $90.5\%$ & 128.7 & 9.30$\times$ & $36.6$ & $42.2$ & $89.0\%$ & 137.9 & 10.01$\times$ & $36.6$ & $51.1$ & $86.7\%$ & 143.1 & 10.38$\times$ \\
\midrule
\multicolumn{19}{l}{\textit{GSM8K (long-reasoning), generation length swept at fixed block length $32$}} \\
\midrule
 128 & 32 &  4 & $66.0$ & $67.0$ & $44.7$ & $65.1\%$ & 77.3 & 2.51$\times$ & $63.0$ & $48.6$ & $62.1\%$ & 51.7 & 1.70$\times$ & $68.0$ & $58.4$ & $54.4\%$ & 45.0 & 1.48$\times$ \\
 256 & 32 &  8 & $66.0$ & $64.0$ & $62.5$ & $75.6\%$ & 67.5 & 3.55$\times$ & $66.0$ & $68.0$ & $73.5\%$ & 75.1 & 3.95$\times$ & $66.0$ & $80.4$ & $68.6\%$ & 63.6 & 3.35$\times$ \\
 512 & 32 & 16 & $72.0$ & $72.0$ & $81.8$ & $84.0\%$ & 75.8 & 5.84$\times$ & $75.0$ & $88.0$ & $82.8\%$ & 108.2 & 8.34$\times$ & $73.0$ & $101.9$ & $80.1\%$ & 92.4 & 7.13$\times$ \\
\midrule
\multicolumn{19}{l}{\textit{GSM8K (long-reasoning), $L$ and $B$ scaled together at fixed block count $8$, so $B/L$ is held constant}} \\
\midrule
 128 & 16 &  8 & $65.0$ & $68.0$ & $47.2$ & $63.1\%$ & 73.2 & 2.39$\times$ & $66.0$ & $52.4$ & $59.0\%$ & 54.5 & 1.79$\times$ & $66.0$ & $60.8$ & $52.5\%$ & 44.8 & 1.47$\times$ \\
 256 & 32 &  8 & $66.0$ & $64.0$ & $62.5$ & $75.6\%$ & 67.5 & 3.55$\times$ & $66.0$ & $68.0$ & $73.5\%$ & 75.1 & 3.95$\times$ & $66.0$ & $80.4$ & $68.6\%$ & 63.6 & 3.35$\times$ \\
 512 & 64 &  8 & $70.0$ & $77.0$ & $76.8$ & $85.0\%$ & 77.2 & 5.94$\times$ & $72.0$ & $80.3$ & $84.3\%$ & 115.4 & 8.90$\times$ & $74.0$ & $97.0$ & $81.1\%$ & 99.8 & 7.69$\times$ \\
\midrule
\multicolumn{19}{l}{\textit{MMLU (short-answer), generation length swept at fixed block length $16$}} \\
\midrule
  16 & 16 &  1 & $56.5$ & $58.0$ & $4.0$ & $75.3\%$ & 128.3 & 3.83$\times$ & $55.5$ & $4.5$ & $72.1\%$ & 82.9 & 2.80$\times$ & $55.0$ & $6.2$ & $61.4\%$ & 59.3 & 1.96$\times$ \\
  32 & 16 &  2 & $61.5$ & $62.5$ & $4.0$ & $87.5\%$ & 247.2 & 7.37$\times$ & $59.0$ & $4.6$ & $85.6\%$ & 147.9 & 4.94$\times$ & $59.5$ & $7.5$ & $76.7\%$ & 93.3 & 3.12$\times$ \\
  64 & 16 &  4 & $64.0$ & $64.0$ & $4.8$ & $92.5\%$ & 490.4 & 16.13$\times$ & $60.0$ & $5.1$ & $92.1\%$ & 355.5 & 11.69$\times$ & $59.5$ & $20.2$ & $68.4\%$ & 77.8 & 2.56$\times$ \\
\bottomrule
\end{tabular}%
}
\caption{\textbf{Generation length and block length.}
Longer generations offer more computation to skip; block length sets the scope of parallel
commitment. DualCache adds per-pass acceleration.
$L$: generation length; $B$: block length; $N{=}L/B$: block count;
Baseline: full-decoding accuracy.}
\label{tab:blocklen}
\end{table}

\noindent\textbf{Longer budgets create more removable computation.}
\label{sec:scaling}
At fixed block length, increasing generation length adds blocks to the reference schedule
(Table~\ref{tab:blocklen}). MMLU uses nearly constant steps under $\mathrm{C}^4$, while GSM8K
steps grow more slowly than the full budget. Savings also increase when generation and block
length grow together at fixed block count. The benefit therefore extends beyond simply
having more blocks to skip: longer within-block budgets also leave work for local acceleration
to remove.

\noindent\textbf{Block length changes how computation is organized.}
At fixed generation length, the HumanEval block-length sweep changes the scope of parallel
updates and how often generation advances to a new block. Step savings vary less than in the
generation-length sweeps, while accuracy changes across configurations. Block length thus
shapes the prediction trajectory and the efficiency of local updates, rather than merely
increasing the total budget available to skip.

\noindent\textbf{Caching changes the cost of the remaining passes.}
Fast-dLLM combines caching with parallel commitment; $\mathrm{C}^4$ additionally skips later
blocks through global stopping, with a pronounced advantage on longer MMLU generations.
DualCache can further accelerate $\mathrm{C}^4$ on longer GSM8K generations, but its benefit
is not uniform across settings. On HumanEval, cached $\mathrm{C}^4$ uses fewer steps than
Fast-dLLM, while uncached $\mathrm{C}^4$ achieves higher accuracy than either cached variant.
Reducing the number of passes and reducing their cost are therefore complementary directions,
but caching also changes the predictions that drive commitment and stopping.

%% file: sections/appendix/fill_quality.tex
\section{Response Quality and Post-Exit Refinement}
\label{app:fill-quality}

The answer and its surrounding explanation can settle at different stages of decoding.
On short-answer tasks, CVEE may reach its stopping condition while parts of the continuation
remain masked. Filling these positions in one pass completes the output, but can introduce
repetition and disrupt fluency. This section examines that behavior on the five short-answer
tasks in Table~\ref{tab:main}, then explores how additional decoding shapes
the completed response.

\paragraph{Quality of the completed response.}
We retain the full text generated by $\mathrm{C}^4$ and full-budget decoding on the same examples.
Table~\ref{tab:fill-quality} characterizes these outputs using generative perplexity under a
frozen GPT-2 (\emph{PPL}), the fraction of outputs that differ from full-budget decoding
(\emph{Changed}), and the fraction of adjacent token repetitions (\emph{Repeat}).

\begin{table}[h]
\centering
\small
\resizebox{\textwidth}{!}{%
\begin{tabular}{lrrrrrrrr}
\toprule
& \multicolumn{4}{c}{LLaDA-8B-Instruct}
& \multicolumn{4}{c}{Dream-7B-Instruct} \\
\cmidrule(lr){2-5}\cmidrule(lr){6-9}
Task
& Base PPL & $\mathrm{C}^4$ PPL & Changed & Repeat
& Base PPL & $\mathrm{C}^4$ PPL & Changed & Repeat \\
\midrule
MMLU        & $27.3$  & $95.3$  & $57\%$ & $19.5\%$
            & $56.5$  & $97.6$  & $22\%$ & $7.0\%$ \\
ARC-C       & $21.4$  & $73.0$  & $86\%$ & $26.9\%$
            & $61.2$  & $81.9$  & $24\%$ & $12.6\%$ \\
HellaSwag   & $44.6$  & $65.1$  & $35\%$ & $3.6\%$
            & $48.4$  & $100.5$ & $49\%$ & $4.7\%$ \\
PIQA        & $37.5$  & $59.9$  & $38\%$ & $2.1\%$
            & $51.3$  & $68.6$  & $26\%$ & $1.4\%$ \\
WinoGrande$^{\dagger}$
            & $518.7$ & $865.2$ & $17\%$ & ---
            & $1216.5$ & $1198.6$ & $2\%$ & --- \\
\bottomrule
\end{tabular}%
}
\caption{\textbf{Full-response quality after one-shot completion.}
PPL is GPT-2 perplexity; Changed measures full-text differences from full-budget decoding;
Repeat measures adjacent token repetition. Early completion often raises perplexity and
introduces repetition in the continuation.
$^{\dagger}$WinoGrande outputs have a median length of two words, limiting the interpretation
of perplexity and repetition.}
\label{tab:fill-quality}
\end{table}

The extracted answer matches full-budget decoding on $98$--$100\%$ of examples, while the
completed text changes more frequently. Quality also varies across tasks with similar exit
times: ARC-C exhibits more repetition than PIQA. The text still unresolved at exit therefore
appears relevant to completion quality alongside the timing of termination.

\paragraph{Allocating computation to refinement.}
After CVEE first fires at step $t$, delayed filling runs ordinary decoding for
$\lceil\alpha\sqrt{t}\rceil$ additional steps before filling the remaining masks.
The parameter $\alpha$ controls the refinement budget, with $\alpha{=}0$ corresponding
to the immediate fill used in the main experiments. These extra updates progressively
resolve the continuation before its final completion.

\begin{table}[h]
\centering
\small
\resizebox{\textwidth}{!}{%
\begin{tabular}{lrrrrrrrrrrrr}
\toprule
& \multicolumn{6}{c}{LLaDA-8B-Instruct}
& \multicolumn{6}{c}{Dream-7B-Instruct} \\
\cmidrule(lr){2-7}\cmidrule(lr){8-13}
& \multicolumn{3}{c}{MMLU}
& \multicolumn{3}{c}{ARC-C}
& \multicolumn{3}{c}{MMLU}
& \multicolumn{3}{c}{ARC-C} \\
\cmidrule(lr){2-4}\cmidrule(lr){5-7}\cmidrule(lr){8-10}\cmidrule(lr){11-13}
$\alpha$
& Acc & Step & Repeat
& Acc & Step & Repeat
& Acc & Step & Repeat
& Acc & Step & Repeat \\
\midrule
Baseline
& $64.0$ & $64.0$ & $0.0\%$
& $86.5$ & $64.0$ & $0.0\%$
& $71.5$ & $64.0$ & $0.0\%$
& $87.5$ & $64.0$ & $0.0\%$ \\
\midrule
\rowcolor{finalres}
$0$
& $64.0$ & $4.8$  & $19.5\%$
& $86.5$ & $3.9$  & $26.9\%$
& $70.0$ & $4.5$  & $7.0\%$
& $88.5$ & $4.2$  & $12.6\%$ \\
$1$
& $63.5$ & $6.9$  & $18.9\%$
& $86.5$ & $5.9$  & $26.7\%$
& $70.0$ & $6.4$  & $7.8\%$
& $88.5$ & $6.1$  & $5.8\%$ \\
$2$
& $63.5$ & $8.4$  & $15.5\%$
& $87.0$ & $7.8$  & $23.0\%$
& $70.5$ & $7.2$  & $5.6\%$
& $88.5$ & $7.1$  & $4.9\%$ \\
$4$
& $63.5$ & $9.9$  & $12.6\%$
& $87.0$ & $9.9$  & $20.5\%$
& $70.5$ & $7.6$  & $5.0\%$
& $88.5$ & $7.4$  & $4.4\%$ \\
$8$
& $63.5$ & $12.7$ & $7.9\%$
& $87.0$ & $14.3$ & $12.0\%$
& $70.5$ & $8.3$  & $2.4\%$
& $88.5$ & $7.7$  & $2.8\%$ \\
$16$
& $63.5$ & $17.8$ & $1.7\%$
& $87.0$ & $22.7$ & $2.4\%$
& $70.5$ & $9.4$  & $0.2\%$
& $88.5$ & $8.2$  & $1.0\%$ \\
\bottomrule
\end{tabular}%
}
\caption{\textbf{Additional refinement trades decoding steps for lower repetition.}
Larger $\alpha$ allocates more updates between the CVEE trigger and the final fill.
Repetition generally falls as this budget grows, with small changes in task accuracy.}
\label{tab:momentum}
\end{table}

The additional updates mainly improve the continuation: repetition falls while task accuracy
changes little. Dream reaches low repetition with fewer added steps than LLaDA, suggesting
that refinement cost depends on the backbone as well as the task. These results identify a
further use for the computation saved by early exit: part of that budget can support response
refinement after the stopping criterion is met. Extending this policy to account for coherence
and completeness would connect the observed reduction in repetition to broader response quality.

%% file: sections/appendix/limitations.tex
\section{Future Directions}
\label{app:limitations}

The interaction between local commitment and global stopping opens several directions for
allocating computation as predictions evolve.

\paragraph{Multimodal diffusion decoding.}
MMaDA \citep{yang2025mmada} extends diffusion modeling to textual reasoning, multimodal
understanding, and text-to-image generation, while LaViDa \citep{li2025lavida} applies diffusion
language modeling to multimodal understanding. MMaDA-Parallel \citep{tian2026mmadaparallel}
further enables bidirectional text--image interaction throughout denoising for thinking-aware
editing and generation. These models offer settings in which to study
local commitment and global stopping beyond text-only inputs. For visually grounded responses,
an extension of CVEE could combine textual agreement with evidence that predictions remain
consistent with the image. The coupled trajectories in MMaDA-Parallel motivate a further question: different
modalities and regions may stabilize at different times, calling for modality-specific commitment
criteria and stopping decisions that account for cross-modal consistency. Evaluating these
criteria would require both task-level correctness and perceptual quality, with calibration
reassessed for the new modalities.

\paragraph{Adaptive response refinement.}
Delayed filling reduces repetition by spending part of the saved budget on the continuation
(Appendix~\ref{app:fill-quality}). This suggests adapting the refinement budget to the amount
and structure of unresolved text. Immediate one-shot filling can leave a repetitive or less fluent
response, so extending refinement to coherence, completeness, and instruction following would
broaden the quality objectives guiding computation.

\paragraph{Revisiting early commitments.}
Later context can provide useful evidence for revising earlier token choices. CCTC currently
retains committed tokens, while CVEE tracks agreement without independently verifying those
tokens. Selective revision could use changes in context to identify positions worth reopening.
Jointly deciding which tokens to revisit and when to stop would extend adaptive computation
to the revision of committed text.

\paragraph{Completion across structured outputs.}
Long programs and multi-part responses require coordination among dependent components.
Extending CVEE's current single-unit tracking to multiple regions could make stopping sensitive
to both local completion and cross-region dependencies. Completed regions could remain fixed
while unresolved components receive further updates, with execution or consistency checks
informing the allocation of computation.

\paragraph{Transfer with statistical risk control.}
The current results establish empirical behavior on the evaluated tasks and backbones;
formal risk guarantees remain an open direction. Statistical risk control
\citep{xie2026statistical,wynn2025controlling} offers a basis for connecting early termination
to explicit error tolerances. Applying it to adaptive commitment would require modeling how
local updates change the observations used for stopping and how calibration transfers under
distribution shift.